\documentclass{article}

\usepackage[main,final]{neurips_2026}

\usepackage[utf8]{inputenc}
\usepackage[T1]{fontenc}
\usepackage{hyperref}
\usepackage{url}
\usepackage{booktabs}
\usepackage{amsfonts}
\usepackage{amsmath}
\usepackage{amssymb}
\usepackage{nicefrac}
\usepackage{microtype}
\usepackage{xcolor}
\usepackage{graphicx}
\usepackage{subcaption}
\usepackage{placeins}
\usepackage{needspace}
\usepackage{enumitem}
\usepackage[capitalize]{cleveref}
\hypersetup{hidelinks,pdfauthor={Moshe Eliasof},
  pdftitle={How Language Models Compress and Compare: Understanding Selection with Token Covariance Maps}}
\ifdefined\XeTeXversion
  \newcommand{\mathbbm}[1]{\mathbf{#1}}
\else
  \usepackage{bbm}
\fi
\graphicspath{{figures/}}

\title{How Language Models Compress and Compare:\\
Understanding  Selection with Token Covariance Maps}

\author{Moshe Eliasof\\
Faculty of Computer and Information Science\\
Ben-Gurion University of the Negev\\
\texttt{eliasof@bgu.ac.il}}

\begin{document}
\maketitle

\begin{abstract}
Many language-model tasks reduce to selecting among structured alternatives, yet we
still lack a simple picture of how a decoder-only transformer organizes an entire
selection prompt across depth. We introduce \emph{token covariance maps}, a prompt-level
geometric view that tracks how token positions co-vary in hidden-state space at each
layer. This turns a structured prompt into a depth-indexed sequence of maps whose
spectral and block structure can be aligned with instructions, context, and candidate
answers.
Across 14 decoder-only models from 4 families and 12 structured-selection benchmarks,
token covariance maps reveal a robust compress-then-compare organization. Early layers
represent prompts diffusely, middle layers compress them into a shared low-rank
bottleneck, and late layers re-expand into context-option and option-option geometry.
This organization recurs across larger fixed-template samples, deterministic prompt
rewrites, pairwise reranking reformulations, and controlled answer-set manipulations,
while also emerging during training from an initially flat depth profile. Its timing
aligns with behavior: useful answer discrimination appears well after the bottleneck, and
low-rank intervention around the bottleneck preferentially disrupts late comparison
geometry.
We additionally show that the same geometric profile appears in
instruction-tuned models, across several languages, and during open-ended
generation. We further show that changing the internal representation of one
candidate around the compressed region can shift which candidate the model
later prefers.
These results show that token covariance maps are a diagnostic
tool for comparing how models organize information across prompts, model
families, and stages of training.
\end{abstract}
\section{Introduction}

Many language-model tasks reduce to choosing among structured alternatives: selecting an
answer in a multiple-choice prompt, comparing two candidates, or reranking a small set of
options. These settings are central to evaluation and are also common in practice, where
pairwise comparison is often an effective prompting strategy for ranking and
selection~\citep{qin2024pairwise}. Yet, we still have only a limited picture of how a
decoder-only transformer organizes the whole prompt while the selection computation
unfolds across depth.

Existing interpretability tools reveal important parts of this process. Attention analyses
show how information is routed between tokens~\citep{clark2019bert,vig2019multiscale};
probing and intermediate-readout methods ask what information or predictions are available
at different layers~\citep{belinkov2022probing,hewitt2019structural,tenney2019bert,belrose2023tunedlens};
representation-similarity methods compare hidden spaces across layers, models, or training
runs~\citep{raghu2017svcca,morcos2018pwcca,kornblith2019cka}; and mechanistic studies
identify circuits, induction heads, MLP memories, or progress measures for specific
behaviors~\citep{elhage2021framework,olsson2022induction,geva2021transformer,wang2023ioi,nanda2023grokking}.
These approaches are powerful, but they usually do not give a single prompt-level object
that shows how instructions, context, and answer candidates are jointly arranged at each
layer.

To bridge this gap, we introduce \emph{token covariance maps}. Given hidden states
$H^{(\ell)} \in \mathbb{R}^{T \times D}$ at layer $\ell$, with $T$ tokens and $D$ hidden
features, we center across token positions and compute a covariance matrix over token
indices. The resulting $T \times T$ map is specific to one prompt and one layer, so it can
be aligned directly with prompt segments such as the context and answer options. Its
spectrum shows whether the prompt representation is spread across many token-position
directions or concentrated into a few dominant modes; its block structure shows how
instructions, context, and answer options interact.

The lens of token covariance maps reveals a simple depth organization for structured
selection: (i) early layers keep the prompt spread across many token-position directions;
(ii) middle layers compress it into a shared low-rank bottleneck; and (iii) late layers
re-expand into structured interactions among the context and candidate answers. We call
this a \emph{compress-then-compare} geometry: the model first compresses the prompt into a
shared mid-layer representation, then builds late geometry for comparing candidate
answers.

We study this geometry across 14 decoder-only models from 4 families and 12
structured-selection benchmarks. The picture is consistent: the low-rank middle stage
appears in the interior of the network, late-layer block structure changes with the type
of selection problem, and the same depth profile persists across prompt reformulations,
answer-set changes, larger fixed-template samples, and training checkpoints. We also
connect the geometry to model behavior: answer preference becomes strongest
only after the compressed middle stage, and perturbing hidden states in that region
changes the late geometry used to compare candidate answers. Token
covariance maps thus expose a recurring layerwise organization of structured
selection: compression first, comparison later.

\textbf{Main contributions.} Our contributions are summarized as follows:
\begin{itemize}[leftmargin=2em]
    \item \textbf{Token covariance maps.}
    We introduce a prompt-level geometric object that turns each layer of hidden states
    into a position-by-position map aligned with the structure of a single prompt.

    \item \textbf{A compress-then-compare depth profile.}
    We show that the same depth profile appears across 14 models,
    4 families, and 12 benchmarks: (i) spread-out early geometry, (ii) a low-rank
    mid-layer bottleneck, and (iii) late re-expansion into answer-comparison structure.

    \item \textbf{Connection to behavior and perturbation.}
    We show that answer preference becomes strongest
    after the compressed middle stage, while perturbing hidden states around that
    stage disrupts late comparison geometry. Our controls connect this region
    to subsequent candidate preference across the compressed region.

    \item \textbf{Robustness across prompts, answer sets, samples, and training.}
    We demonstrate that the same bottleneck is preserved under
    pairwise reranking, deterministic template rewrites, controlled answer-set
    manipulations, and higher-sample fixed-template reruns, and that
    it appears during training rather than being present at initialization.

    \item \textbf{A diagnostic view of structured selection.}
    We use the maps to compare prompts, checkpoints, model families,
    and task types. Comparisons with direct hidden-state rank,
    centered cosine, Gram, CKA, and SVCCA clarify what scalar spectra recover
    and what token-position maps add for candidate-level analysis.
\end{itemize}
\section{Related Work}
\label{sec:related}

\textbf{Geometry and anisotropy in language-model representations.}
Contextual language-model representations are highly anisotropic and layer-dependent~\citep{ethayarajh2019contextual}. Related work shows that representations can degenerate into narrow cones~\citep{gao2019representation}, and that a few rogue dimensions can dominate cosine and Euclidean similarity in transformer language models~\citep{timkey2021bark}. Broader studies of deep representations analyze spectral structure, implicit self-regularization, and collapse phenomena~\citep{martin2021implicit,papyan2020neural}. ALL-IN uses node covariance for cross-dataset graph transfer~\citep{eliasof2026bridging}. Rather than analyze raw token embeddings, vocabulary embeddings, or class-level collapse, we form a covariance map over token positions for a single prompt and layer. The rows and columns are prompt positions, so the geometry can be read in terms of instructions, context, and answer options.

\textbf{Probes, attention, and transformer mechanisms.}
Attention analyses expose token-to-token routing patterns~\citep{clark2019bert,vig2019multiscale}. Probing methods ask which linguistic properties are recoverable from hidden states~\citep{hewitt2019structural,tenney2019bert,belinkov2022probing}. Mechanistic interpretability studies identify component-level structure such as transformer circuit motifs~\citep{elhage2021framework}, induction heads~\citep{olsson2022induction}, feed-forward memories~\citep{geva2021transformer}, and task-specific circuits or progress measures~\citep{wang2023ioi,nanda2023grokking}. Token covariance maps complement these component-level and prediction-level analyses by summarizing the sequence-level geometry produced up to each layer. This makes it possible to study how a whole structured-selection prompt changes through depth.

\textbf{Representation similarity and spectral tools.}
Singular Vector Canonical Correlation Analysis (SVCCA) compares learned representations through canonical-correlation structure across layers or training time~\citep{raghu2017svcca}. Projection-Weighted Canonical Correlation Analysis (PWCCA) weights canonical directions by their contribution to the original representation~\citep{morcos2018pwcca}. Centered Kernel Alignment (CKA) compares representations through normalized kernel similarity and is widely used for layer- and architecture-level comparison~\citep{kornblith2019cka,raghu2021vit}. GLOT uses token-feature similarity graphs as input to graph neural networks to improve global text representations~\citep{mantri2026tokengraphs}. Our object is per-example and position-resolved: for each prompt, we study the spectrum and block structure of a token-position covariance map. Effective rank summarizes compression in this space~\citep{roy2007effective}, while block summaries connect late re-expansion to context-option and option-option structure.

\textbf{Layerwise prediction and adaptive computation.}
The logit lens and tuned lens decode intermediate transformer states into vocabulary predictions, giving a layerwise view of prediction evolution~\citep{belrose2023tunedlens}. Early-exit and adaptive-computation methods attach intermediate exits or confidence criteria to reduce inference cost, including DeeBERT for encoder models~\citep{xin2020deebert} and Confident Adaptive Language Modeling (CALM) for language modeling~\citep{schuster2022calm}. We use intermediate readouts to test when useful answer discrimination appears relative to the compressed middle stage, as shown in \Cref{sec:timing}.

\textbf{Bottleneck terminology.}
\citet{kong2025token} advocate task-aware token reduction as a design principle for improving generative models beyond computational efficiency. The information bottleneck method formalizes compression as preserving information about a target variable under an information-theoretic tradeoff~\citep{tishby2000ib}. We distinguish the geometric bottleneck studied here from an information-bottleneck training objective. The interventions in \Cref{sec:functional} test how the compressed region contributes to late comparison geometry. Our bottleneck is an effective-rank minimum at fixed sequence length, as shown in \Cref{fig:phase_transition}.

\textbf{Reranking, pairwise comparison, and selective reformulation.}
Pairwise ranking prompting has been used to make language models compare candidate items more effectively~\citep{qin2024pairwise}. We use pairwise reformulation as a test of geometric preservation: if token covariance maps reflect compare-and-select computation rather than one multiple-choice template, the bottleneck should persist when a multi-option prompt is rewritten as candidate-A versus candidate-B comparisons. The selective-reranking benchmark in Appendix~\ref{sec:appendix_side_analyses} separately tests whether covariance features add value beyond confidence for deciding when to rerank.

\section{Token Covariance Maps}
\label{sec:method}

This section introduces the geometric objects used throughout the paper. We first define
token covariance maps in \Cref{sec:covariance_definition}: per-prompt, per-layer maps
whose rows and columns correspond to token positions. We then define spectral summaries
for tracking compression across depth in \Cref{sec:spectral_summaries}, block summaries
for reading context-option and option-option structure in
\Cref{sec:block_summaries}, behavioral readouts in \Cref{sec:behavioral_readouts}, and
the low-rank perturbation used to probe the bottleneck region in
\Cref{sec:low_rank_perturbation}. These quantities are used together in
\Cref{sec:experiments}: the spectral summaries establish the compress-then-compare
trajectory, the block summaries characterize late comparison geometry, the behavioral
readouts align this geometry with answer selection, and the perturbation test evaluates
the functional role of the compressed middle stage.

\subsection{From Hidden States to Token Covariance Maps}
\label{sec:covariance_definition}

Let $H^{(\ell)} \in \mathbb{R}^{T \times D}$ denote the hidden states at layer $\ell$,
where $T$ is the sequence length and $D$ is the hidden dimension. We center across token
positions,
$\tilde{H}^{(\ell)} = H^{(\ell)} - \frac{1}{T}\mathbf{1}\mathbf{1}^{\top}H^{(\ell)},$
and define the token covariance map
\begin{equation}
C^{(\ell)} =
\frac{1}{D}\tilde{H}^{(\ell)}(\tilde{H}^{(\ell)})^\top
\in \mathbb{R}^{T \times T}.
\end{equation}
Entry $C^{(\ell)}_{ij}$ measures covariance between token positions $i$ and $j$ in
hidden-feature space. The map summarizes the symmetric association between token representations, complementing attention-based routing analyses. It is a
prompt-specific summary of the hidden-state geometry produced by the network up to layer
$\ell$, and its rows and columns can be aligned directly with prompt segments. We visualize
these maps directly in \Cref{sec:law} and use them as the base object for all subsequent
spectral, block-level, behavioral, and perturbation analyses.

With the same centering, centered Gram equals $DC^{(\ell)}$
and has identical scale-invariant summaries. Centered cosine additionally
normalizes each token vector. Appendix~\ref{sec:extended_methods} details
these conventions.

\subsection{Tracking Compression Across Depth}
\label{sec:spectral_summaries}

We use spectral summaries of $C^{(\ell)}$ to track compression in token-position space.
Let $\lambda_1 \ge \cdots \ge \lambda_T \ge 0$ be the eigenvalues of $C^{(\ell)}$, and
define
$p_i = \frac{\lambda_i}{\sum_j \lambda_j}.$
Our main compression statistic is the effective rank~\citep{roy2007effective},
\begin{equation}
\operatorname{erank}(C^{(\ell)}) =
\exp\!\left(-\sum_i p_i \log p_i\right).
\end{equation}
High effective rank means that covariance mass is spread across many token-position
directions; low effective rank means that it is concentrated in a few dominant modes. We
define the bottleneck depth as the relative depth where effective rank reaches its
minimum along a sampled layer trajectory. This is the main statistic used in
\Cref{sec:law} to identify the compress-then-compare depth profile and in
\Cref{sec:stress} to test whether the same bottleneck is preserved under prompt and
answer-set changes.

We also track the top-1 eigenvalue fraction,
$\rho_1(C^{(\ell)}) = \frac{\lambda_1}{\sum_j \lambda_j},$
and the band-4 energy fraction,
\begin{equation}
E_{\mathrm{band}}(C^{(\ell)}) =
\frac{\sum_{|i-j|\le 4}|C^{(\ell)}_{ij}|}
{\sum_{i,j}|C^{(\ell)}_{ij}|}.
\end{equation}
The top-1 fraction measures dominance of the leading covariance direction, while band
energy measures how much covariance mass remains near the token diagonal. Together with
effective rank, these diagnostics describe the compression and re-expansion trajectories
reported in \Cref{sec:law}.

\subsection{Reading Comparison Structure from Prompt Blocks}
\label{sec:block_summaries}

Structured-selection prompts have natural parts: instructions, context, answer options,
and an answer position. Given token index sets
$
B_{\mathrm{inst}},\;
B_{\mathrm{ctx}},\;
B_{\mathrm{opt}_1},\ldots,B_{\mathrm{opt}_K},\;
B_{\mathrm{ans}},
$
we define the average absolute covariance mass between blocks $A$ and $B$ at layer
$\ell$ as
\begin{equation}
S_\ell(A,B)=
\frac{1}{|A||B|}
\sum_{i\in A}\sum_{j\in B}|C^{(\ell)}_{ij}|.
\end{equation}
This yields three task-aligned summaries:
\begin{align}
\mathrm{CtxOpt}^{(\ell)}
&=
\frac{1}{K}\sum_{m=1}^{K}
S_\ell(B_{\mathrm{ctx}},B_{\mathrm{opt}_m}), \\
\mathrm{OptSelf}^{(\ell)}
&=
\frac{1}{K}\sum_{m=1}^{K}
S_\ell(B_{\mathrm{opt}_m},B_{\mathrm{opt}_m}), \\
\mathrm{OptCross}^{(\ell)}
&=
\frac{1}{K(K-1)}
\sum_{m\ne m'}
S_\ell(B_{\mathrm{opt}_m},B_{\mathrm{opt}_{m'}}).
\end{align}
Here $\mathrm{CtxOpt}$ measures context-option coupling, $\mathrm{OptSelf}$ measures
within-option structure, and $\mathrm{OptCross}$ measures interaction among distinct
answer options. These summaries test whether late-layer re-expansion follows the structure of the selection problem. We
use them in \Cref{sec:comparison} to characterize late comparison geometry and in
\Cref{sec:stress} to test how late geometry changes when the answer set is manipulated.

For a single scalar summary of final-layer answer-comparison geometry, we use
$R_{\mathrm{cross/self}} = \frac{\mathrm{OptCross}^{(L)}}{\mathrm{OptSelf}^{(L)}},$
where $L$ is the final sampled layer. Larger values indicate stronger inter-option
interaction relative to within-option structure. This ratio is also the main geometric
quantity used in the perturbation experiment of \Cref{sec:intervention} and in the
diagnostic comparisons of \Cref{sec:why_covariance}.

\subsection{Behavioral Readouts}
\label{sec:behavioral_readouts}

For behavioral comparisons, each answer option is scored by its mean next-token
log-probability under the model. For a prompt $x$, let option
$a_m$ contain tokens $(y_{m,1},\ldots,y_{m,n_m})$, and let $x_m$ denote the
prefix preceding that option. We define
\begin{equation}
s_m(x)=
\frac{1}{n_m}\sum_{t=1}^{n_m}
\log p_\theta\!\left(y_{m,t}\mid x_m, y_{m,<t}\right).
\end{equation}
The model prediction is
$\hat{m}(x)=\arg\max_m s_m(x),$
and the confidence margin is the gap between the top two option scores,
$\gamma(x)=s_{\hat{m}(x)}(x)-\max_{m\ne \hat{m}(x)} s_m(x).$
Correctness is the indicator $\mathbbm{1}\{\hat{m}(x)=m^\star\}$, where $m^\star$ is the
gold option. We use the same scoring rule for multiple-choice evaluation,
intermediate-layer readouts, and reranking comparisons. In \Cref{sec:timing}, the
intermediate-layer readouts test when answer discrimination appears relative to the
covariance bottleneck; in \Cref{sec:stress}, the same scoring rule evaluates direct and
reranked prompts.

Appendix~\ref{sec:extended_methods} specifies the scoring protocols
used in the intervention studies.

\subsection{Low-Rank Perturbation Around the Bottleneck}
\label{sec:low_rank_perturbation}

To test whether the compressed middle stage contributes to late comparison geometry, we
perturb hidden states at a target layer $\ell^\star$. Let
$
C^{(\ell^\star)} = U\Lambda U^\top
$
be the eigendecomposition of the token covariance map at that layer. For retained-rank
fraction $\alpha \in (0,1]$, we set
$
r = \max\{1,\lceil \alpha T \rceil\},
$
retain the first $r$ token-space eigenvectors $U_r$, form $P_r=U_rU_r^\top$, and replace
the layer output with
\begin{equation}
\hat{H}^{(\ell^\star)} = P_r H^{(\ell^\star)}.
\end{equation}
This keeps only the dominant token-position covariance modes at the target layer. In
\Cref{sec:intervention}, we compare perturbations at the empirical bottleneck with
matched control layers at offsets $-3$ and $+3$, using $\alpha \in \{0.25,0.50\}$.

The perturbation metric is the change in final-layer comparison geometry,
\begin{equation}
\Delta R_{\mathrm{cross/self}}(\ell^\star,\alpha)=
R_{\mathrm{cross/self,baseline}} -
R_{\mathrm{cross/self,intervened}}.
\end{equation}
Larger values indicate greater disruption of late inter-option comparison structure. We
use these comparisons to assess the contribution of the compressed
region to constructing late comparison geometry, as shown in \Cref{sec:intervention}.

Random-projection and Gaussian controls match the achieved
Frobenius perturbation of low-rank truncation for each prompt and layer, as specified in
Appendix~\ref{sec:extended_methods}.

\section{Experiments}
\label{sec:experiments}

We study four questions: \textbf{(Q1)} whether compression and re-expansion
appear across models, tasks, and sample sizes; \textbf{(Q2)} how the late
geometry relates to candidate comparison; \textbf{(Q3)} when answer
discrimination emerges and how interventions affect it; and \textbf{(Q4)}
whether the profile persists across prompt reformulations, answer-set changes,
and training. Appendix~\ref{sec:appendix_protocols} gives the additional
protocols, while Appendices~\ref{sec:appendix_geometry}--\ref{sec:appendix_side_analyses}
report the complete controls and diagnostic results.

\subsection{Experimental Setup}
\label{sec:setup}

\textbf{Models and tasks.}
We study {14} decoder-only language models from {4} open-weight families:
Pythia, SmolLM, Qwen, and Mistral. The canonical analysis uses {12}
structured-selection benchmarks: PIQA, BoolQ, HellaSwag, ARC-Easy, ARC-Challenge, COPA,
WinoGrande, OpenBookQA, CommonsenseQA, Social IQa, SciQ, and LogiQA.

\textbf{Protocol.}
Unless otherwise stated, datasets use a unified multiple-choice prompt template with 120
validation examples per dataset-model pair. Aggregate depth trajectories use five
relative depth locations $\{0.0,0.25,0.5,0.75,1.0\}$; intermediate readouts use nine.
Additional evaluations vary the prompt format, answer set, sample size, reranking
formulation, training checkpoint, and perturbation type, with full details and figures in
the appendix.

The canonical analysis contains 19,880 prompt--model observations,
with COPA capped at 100 examples per model. The following experiments state their
sample counts. Bracketed intervals denote bootstrap 95\% confidence intervals
unless stated otherwise.

\subsection{A compress-then-compare law across models and tasks}
\label{sec:law}

\begin{figure}[t]
\centering
\includegraphics[width=\textwidth]{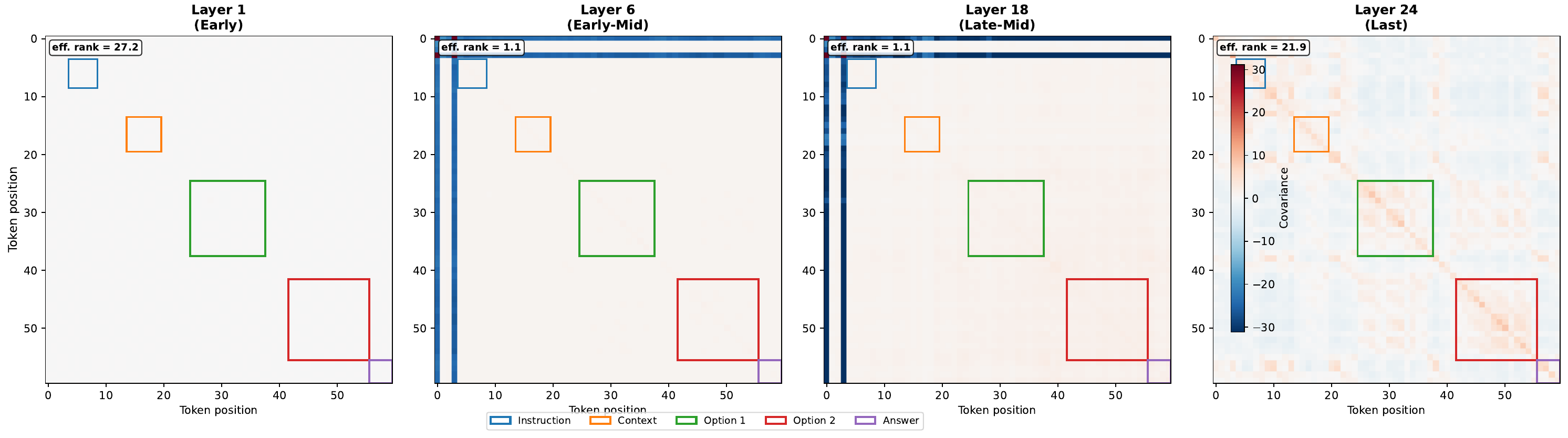}
\caption{\textbf{Token covariance maps across depth for one PIQA prompt in Pythia-410M.} Colored rectangles mark prompt segments. The key visual trajectory is broad early geometry, a low-rank mid-layer bottleneck, and late recovery of off-diagonal context-option and option-option interactions.}
\label{fig:heatmap}
\end{figure}

This subsection answers \textbf{(Q1)} by testing whether the depth profile is visible in individual
prompts and stable in aggregate. Figure~\ref{fig:heatmap} introduces the core object on a
single PIQA prompt. The trajectory is already visible before averaging: early layers have
broadly distributed covariance, middle layers collapse into a narrow low-rank structure,
and late layers recover off-diagonal interactions among context and answer-option regions.

\begin{figure}[t]
\centering
\includegraphics[width=\textwidth]{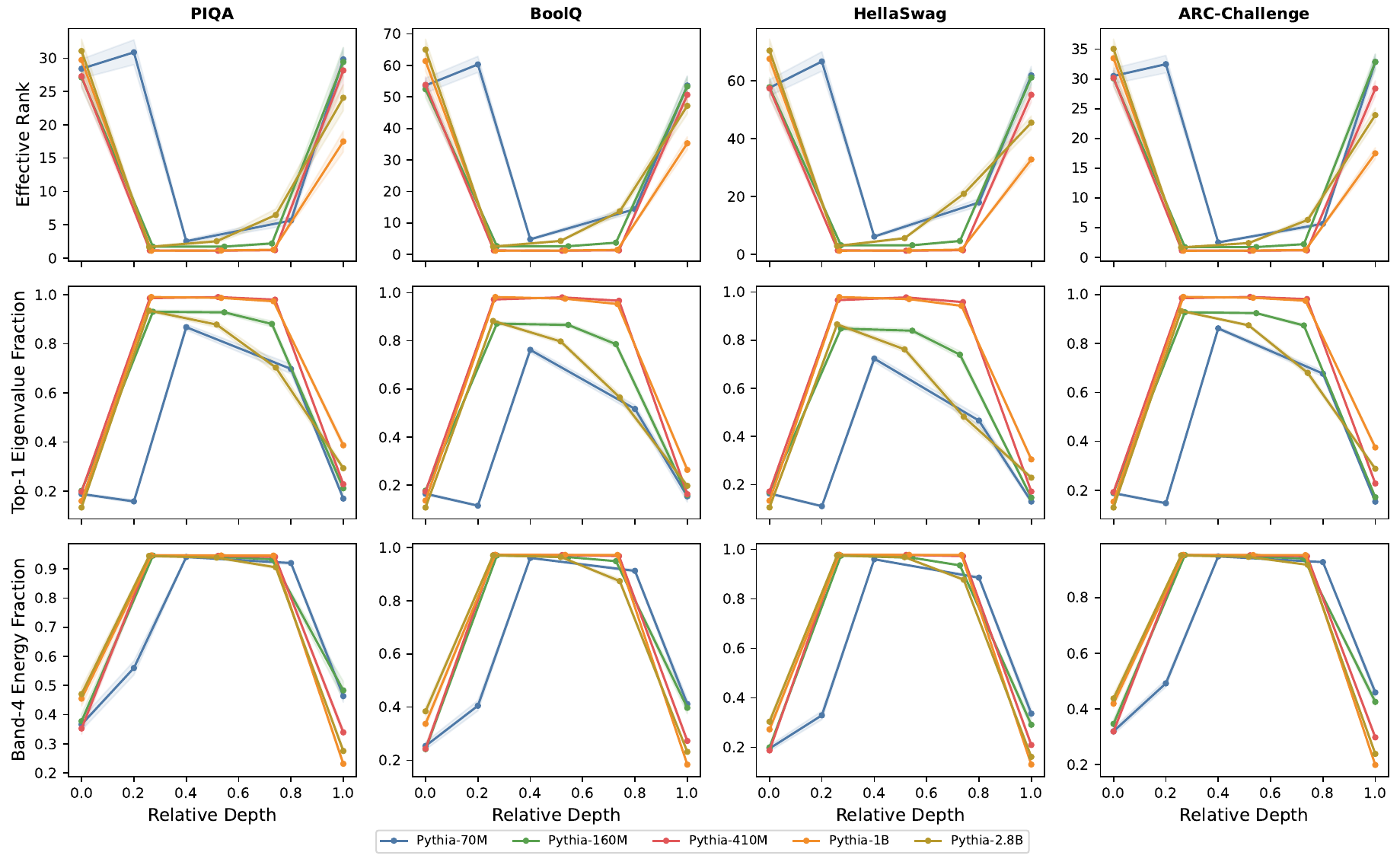}
\caption{\textbf{Robust compression and re-expansion on representative datasets.} Rows show effective rank, top-1 eigenvalue fraction, and band-4 energy; columns show datasets; curves show Pythia scales. The same mid-layer compression and late re-expansion appear in aggregate trajectories, not only in single-prompt heatmaps.}
\label{fig:phase_transition}
\end{figure}

The same trajectory appears after aggregation, as shown in
Figure~\ref{fig:phase_transition}. Effective rank falls to an interior minimum
and then rises, while the top-1 eigenvalue fraction increases during compression
and band energy peaks before longer-range structure reappears. All 14 models
from the four families retain this profile. The cross-family comparison in
Appendix Figure~\ref{fig:cross_family} tests its consistency across architectures;
Appendix~\ref{sec:appendix_visualizations} provides additional prompt-level maps.

We repeat the fixed-template analysis with Pythia-2.8B,
Qwen2.5-7B, and Mistral-7B-v0.3 using 299--500 examples per dataset.
All settings retain an interior bottleneck and late re-expansion, as shown
in Appendix Figure~\ref{fig:high_sample_main_law}. Every-layer rewrites and
candidate edits also preserve the minimum and its surrounding compressed
region, as reported in Appendix Table~\ref{tab:full_layer}.

\subsection{Late re-expansion builds answer-comparison geometry}
\label{sec:comparison}

\begin{figure}[t]
\centering
\includegraphics[width=\textwidth]{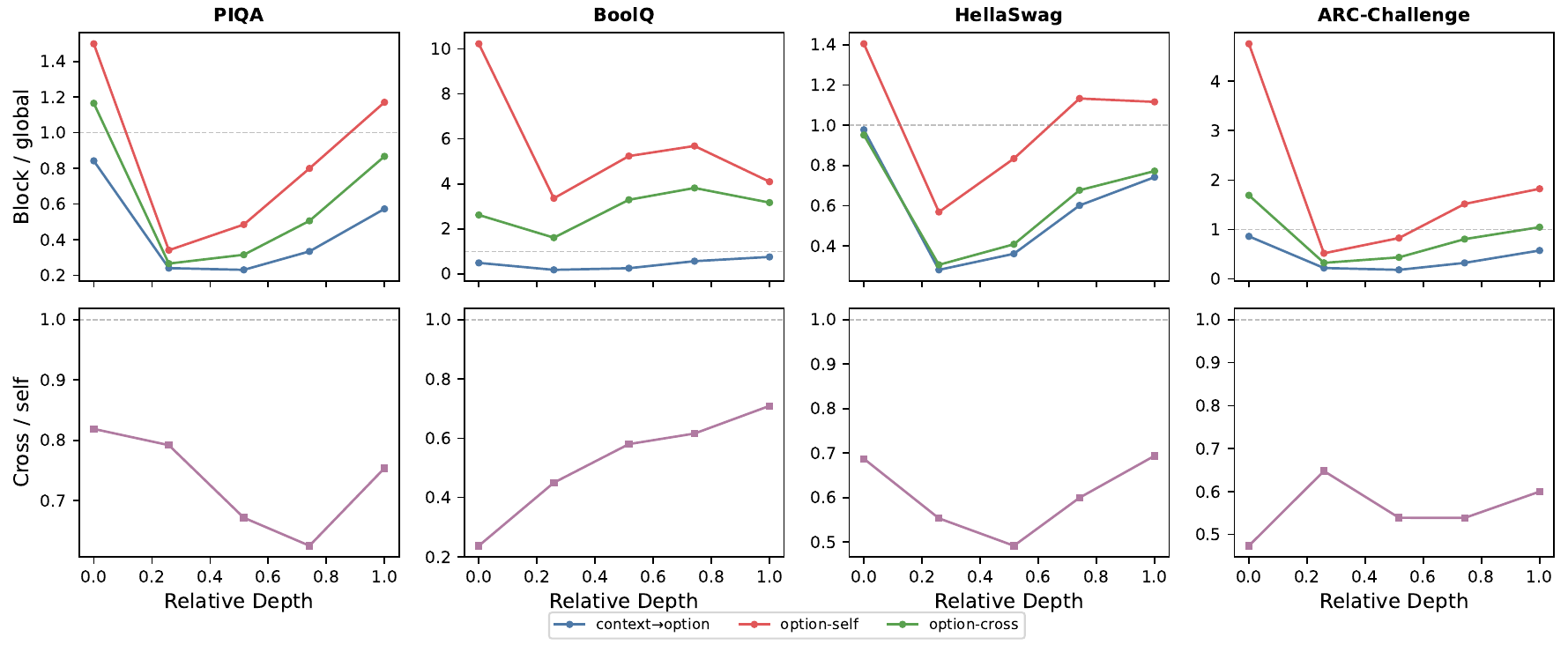}
\caption{\textbf{Late-layer block structure aligns with answer comparison.} Curves for Pythia-2.8B summarize context-option coupling, option-self structure, option-option interaction, and the final cross/self ratio. Late-layer block interactions vary with task type.}
\label{fig:block_structure}
\end{figure}

\textbf{(Q2)} asks what is built after compression. If the late rise in effective rank were only a
generic recovery of dimensionality, it would not need to respect the structure of the
selection problem. Instead, Figure~\ref{fig:block_structure} shows that late-layer block
summaries align with the prompt parts used for answering: binary QA emphasizes
context-option coupling, procedural tasks emphasize option-self structure, and knowledge
or completion tasks show stronger option-cross interaction. The post-bottleneck
re-expansion is therefore organized around answer comparison. This is the main reason we
refer to the late phase as comparison rather than merely recovery of rank: the recovered
structure follows the semantic partition of the selection prompt.

\par\begingroup
To separate instruction wording from candidate content, we compare selection
and neutral-inspection prompts with identical non-instruction tokens, using
Pythia-2.8B, Qwen2.5-7B, and Mistral-7B-v0.3 on PIQA, BoolQ, HellaSwag,
and ARC-Challenge. Their compressed regions nearly coincide. Replacing
candidates with approximately length-matched placeholders retains compression but reduces
late block coupling, as reported in Appendix Table~\ref{tab:objectives}.

We also examine signed correct-versus-wrong blocks, as shown in Appendix
Table~\ref{tab:signed}. Reversing a relation changes late signed
geometry more than renaming entities under exact token-count matching in both
Pythia-2.8B and Qwen2.5-7B, without shifting the bottleneck.
Appendix~\ref{sec:semantics} gives the controls, including the
duplicate-distractor boundary where geometry and accuracy need not move together.
\par\endgroup

\subsection{Behavior and perturbation localize the compressed stage}
\label{sec:functional}
\label{sec:timing}
\label{sec:intervention}

\begin{figure}[t]
\centering
\includegraphics[width=\textwidth]{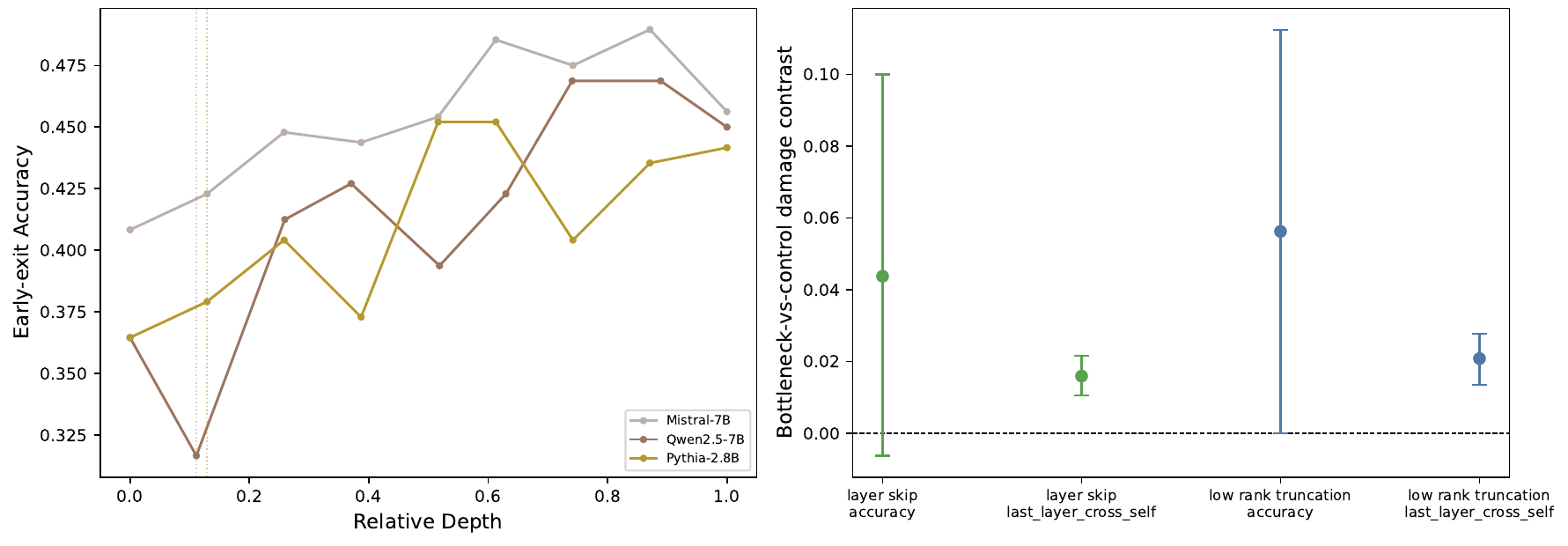}
\caption{\textbf{Functional timing and bottleneck-region perturbation.} Intermediate accuracy peaks after the bottleneck, and low-rank truncation near the bottleneck region preferentially disrupts late comparison geometry. The perturbation tests the functional importance of the compressed region.}
\label{fig:functional_timing}
\end{figure}

\textbf{(Q3)} connects these geometric stages to answer selection. We evaluate
intermediate readouts at nine relative depths for Pythia-2.8B, Qwen2.5-7B,
and Mistral-7B-v0.3. As shown in Figure~\ref{fig:functional_timing}, peak
intermediate accuracy occurs well after the bottleneck in all three models.
For example, Qwen2.5-7B reaches its peak near relative depth 0.74, compared
with a bottleneck near 0.11. The compressed stage therefore precedes the
layers at which the model most clearly prefers the correct answer.

We also perturb hidden states around the empirical bottleneck. Across Pythia-2.8B and
Qwen2.5-7B on PIQA, HellaSwag, and ARC-Challenge, low-rank truncation at the bottleneck
disrupts final-layer cross/self comparison structure more than matched control
layers at offsets $-3$ and $+3$. Figure~\ref{fig:functional_timing} summarizes the timing
and perturbation results. Dense layer sweeps and distortion-matched controls in
Appendix Figures~\ref{fig:dense_mech} and~\ref{fig:distortion_mech} keep the mechanistic
interpretation regional: the compressed middle stage contributes to late comparison
geometry. These comparisons resolve effects across the compressed
region; attribution to individual layers and covariance directions remains open.

\par\begingroup
\paragraph{Do candidate states change subsequent preference?}
To test whether candidate states affect subsequent selection, we construct
pairs differing in one exactly token-length-matched candidate, fixing its
position and every token outside the span. We copy only the replacement
candidate residual states into the original pass at a selected layer,
then score its full text alongside the original candidates.

Figure~\ref{fig:candidate_patching} shows the results for Pythia-2.8B
and Qwen2.5-7B on PIQA, HellaSwag, and ARC-Challenge. At the bottleneck, the
replacement gains preference and becomes the top prediction on approximately
one in ten examples. The margin difference between candidate-state patching
and its control has a positive
95\% confidence interval in 70 of 72 prespecified comparisons, including equal-size position
patches and norm-matched perturbations. Appendix~\ref{sec:patching_details}
reports the held-out evaluation protocol, effect estimates, and all controls.
\begin{figure}[t]
\centering
\includegraphics[width=\textwidth]{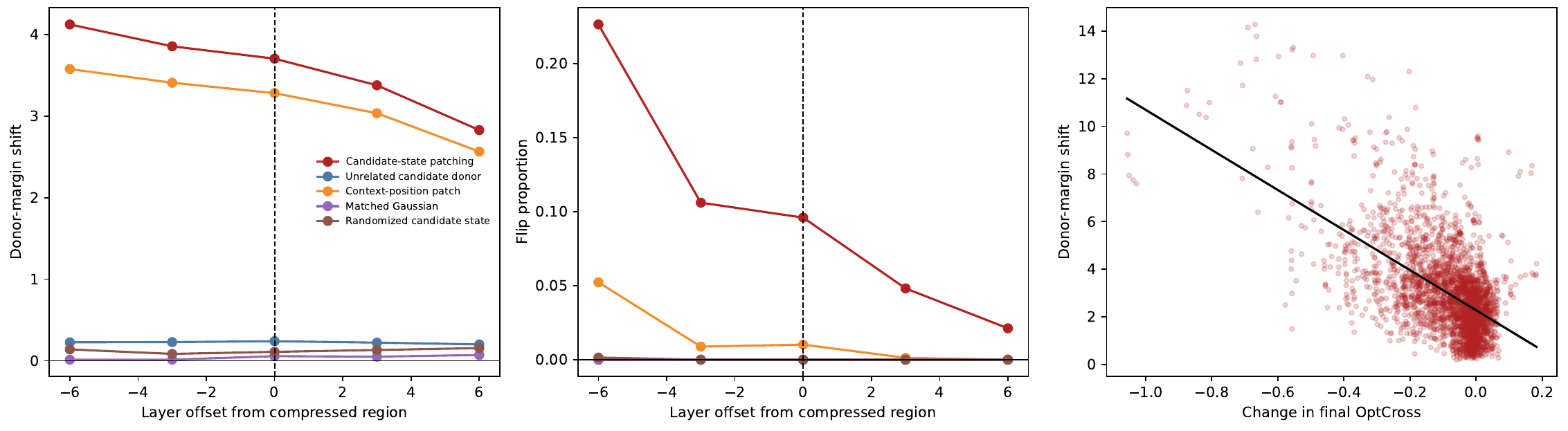}
\caption{\textbf{Changing candidate states changes later candidate preference.}
Pythia-2.8B and Qwen2.5-7B on PIQA, HellaSwag, and ARC-Challenge.
Only the states of the changed candidate are copied between token-length-matched
prompts. Panels show donor-margin change, directed prediction switches, and
geometry--behavior association. The donor is the replacement candidate; scores
use mean full-text log probability. Results use held-out examples.}
\label{fig:candidate_patching}
\end{figure}

At the bottleneck, attention raises effective rank and the MLP lowers it in
all six settings, as shown in Appendix Table~\ref{tab:sublayers}.
\par\endgroup

\subsection{The geometry persists across reformulation, answer-set changes, and training}
\label{sec:stress}

\textbf{(Q4)} examines changes to prompt format, candidate content, and
training stage. Pairwise reranking preserves bottleneck depth and minimum
rank, as shown in Appendix Figure~\ref{fig:reranking}. Deterministic
rewrites and label/order permutations retain the bottleneck within one
sampled depth, with effective-rank trajectory correlations above 0.999,
as shown in Appendix Figure~\ref{fig:prompt_robustness}.

Answer-set manipulations separate bottleneck stability from late comparison
sensitivity. With Pythia-2.8B and Qwen2.5-7B on PIQA, HellaSwag, and
ARC-Challenge, the bottleneck remains fixed while candidate replacements
change the late block summaries. For example, replacing a wrong option with
an irrelevant distractor reduces final cross/self by about 0.11. Appendix
Figure~\ref{fig:answer_set_manipulation} reports the other edits, including
the duplicate-wrong case in which accuracy changes more than the pooled
geometry.

Finally, the bottleneck is learned. In Pythia checkpoints, the effective-rank profile is
nearly flat at initialization, then develops a mid-layer bottleneck over training,
as illustrated in Appendix Figure~\ref{fig:training}. Thus the same compress-then-compare organization is
stable to surface reformulations and candidate-set changes, but is not present before
training. This combination rules out the simplest explanations based only on one prompt
template, one answer order, or an untrained architectural prior.

\par\begingroup
\subsection{Generalization across Formats, Languages, and Generation}
\label{sec:generalization}
We further examine whether the depth profile extends beyond base-model English
prompts. It remains present with official chat templates, aligned XCOPA examples
in five languages, and professional-domain questions, as reported in Appendix
Table~\ref{tab:generalization}.

We also follow the representation during generation. Pythia-2.8B,
Qwen2.5-7B, and Mistral-7B-v0.3 generate option-removed responses on PIQA,
BoolQ, HellaSwag, and ARC-Challenge; Qwen and Mistral additionally use GSM8K,
TriviaQA, MBPP, and QASPER. Compression and re-expansion persist as successive
response prefixes are replayed through every layer. Ordered continuations
produce stronger continuation--answer coupling than token and local-block
shuffles, as shown in Appendix Figure~\ref{fig:online}.
Appendix~\ref{sec:generalization_details} gives the protocols.
The generation experiments extend the observed geometric profile, with
answer quality considered separately from re-expansion.
\par\endgroup

\subsection{Comparison with existing representation analyses}
\label{sec:why_covariance}

\par\begingroup
Under a matched four-feature protocol, direct hidden-state rank gives the
strongest capability fit and centered cosine the best task-family transfer,
as shown in Appendix Table~\ref{tab:baselines}. Covariance,
uncentered Gram, CKA, and SVCCA are evaluated on the same benchmark,
holding out complete models.

Direct hidden-state spectra also recover the bottleneck. Token-position maps
additionally retain candidate coordinates and, for covariance, signed
association. The spectrum and candidate-level structure are thus available
in one object. Appendix Figure~\ref{fig:checkpoint_ranking} reports the
complementary model-ranking diagnostic.
\par\endgroup

\clearpage
\section{Discussion and Conclusion}
\label{sec:discussion_conclusion}

Token covariance maps reveal a recurring layerwise organization for structured selection
in decoder-only language models. Across {14 models}, {4 families}, and
{12 datasets}, structured prompts move through three geometric phases: early
representations are spread across many token-position directions, middle layers compress
the prompt into a low-rank shared bottleneck, and late layers re-expand into
answer-comparison structure. This profile persists under higher-sample reruns,
deterministic prompt rewrites, pairwise reranking reformulations, and controlled
answer-set manipulations, and it develops during training rather than appearing at
initialization.

The late phase gives the bottleneck its computational interpretation. Re-expansion is not
generic rank recovery: its block structure tracks context-option and option-option
interactions, and these interactions vary with the type of selection problem. Behavioral
readouts place answer discrimination after the compressed middle stage, while low-rank
perturbations around that stage disrupt the final comparison geometry.
The interventions identify functional contributions across the
compressed region, as shown in \Cref{sec:functional}.

Token covariance maps support offline comparisons of prompts,
training checkpoints, and model families. Their candidate-resolved summaries
complement scalar spectra. Appendix~\ref{sec:appendix_side_analyses} reports
the selective-reranking and routing comparisons with confidence-based
baselines at matched budgets.

\textbf{Limitations.}
Position, formatting, tokenization, option length, and hidden-state norms
affect magnitudes more than bottleneck location, as examined in
Appendix~\ref{sec:normalization}. Covariance measures symmetric
association; causal direction and a complete semantic account require
additional analyses. Interventions establish regional contributions, while
layer-specific and direction-specific attribution remains open. Generation
extends the observed geometric profile, with answer quality evaluated separately.

The main conclusion is that structured selection in decoder-only language models has a
recurring geometric signature. The model first compresses the prompt into a shared
mid-layer representation, then constructs late geometry for comparing candidate answers.
Token covariance maps make this computation visible at the level of whole prompts, giving
a compact way to study how language models organize selection problems across depth.

\clearpage
\newpage
\bibliographystyle{plainnat}
\bibliography{references}

\clearpage
\newpage
\appendix
\Needspace{8\baselineskip}
\section{Experimental Protocols}
\label{sec:appendix_protocols}
This appendix gives the additional protocols for the main experiments.
Appendix~\ref{sec:appendix_geometry} covers the geometry and robustness checks,
Appendix~\ref{sec:appendix_content} examines candidate content,
and Appendix~\ref{sec:appendix_mechanistic} reports behavioral and mechanistic
controls. Representation comparisons appear in Appendix~\ref{sec:appendix_diagnostics},
generalization in Appendix~\ref{sec:appendix_generalization},
and practical diagnostics in Appendix~\ref{sec:appendix_side_analyses}.
\par\begingroup
\subsection{Additional Control Protocols}
\label{sec:extended_methods}
The definitions in Section~\ref{sec:method} are used throughout the paper.
Here we specify the normalization, compressed-region agreement,
candidate-state transfer, and achieved-distortion matching used in the
additional controls.
\par\begingroup
\paragraph{Normalization, interval agreement, and perturbation controls.}
\label{sec:methods}

Centering specifies the geometric object being measured. With the token-centering
operator $J=I-\mathbf{1}\mathbf{1}^{\top}/T$, the centered token Gram matrix is
$G_c=(JH)(JH)^\top=DC$. Consequently, centered Gram and token covariance have
identical scale-invariant spectral summaries. Multiplying a map by a positive
scalar preserves effective rank, the minimizing layer, and cross/self. Centered
cosine instead divides each centered token vector by its own norm before taking
inner products, and can therefore change these quantities.

For comparisons at every transformer layer, let $r_\ell$ denote effective rank
and $r_{\min}=\min_\ell r_\ell$. We define the compressed region as the connected
layer interval containing the minimum for which
$r_\ell\le r_{\min}+0.1(r_{\mathrm{early}}-r_{\min})$.
The intersection-over-union (IoU) of two regions is the number of layers in their
intersection divided by the number in their union. This measures agreement over
an interval rather than only agreement between two minimizing layers.

Layer interventions and candidate-state patching use
a model-level reference estimated from the canonical rank minima, rather
than a separate every-layer minimum for each prompt. The intervention is
applied to the output of the next transformer block after the reference
hidden state. Offset zero denotes this intervention layer; the localization
claim concerns the compressed region, not the exact minimizing layer.

\paragraph{Candidate-state transfer.}
A paired prompt replaces exactly one candidate with a text of the same token
length, retaining its position and every token outside that span. We call the
replacement the donor candidate. At a selected layer, we copy only its residual
states into the original prompt and continue the forward pass. The evaluation set
contains the original candidate texts and the donor text. The donor margin is its
mean continuation log probability relative to the largest score among the original
candidates; positive values favor the donor. A directed switch occurs when the donor becomes the highest-scoring
candidate after patching but was not highest-scoring before patching. These
quantities test candidate preference using full texts rather than answer-label
tokens. \Cref{sec:patching_details} gives the controls, split, and matching
procedure.

\paragraph{Scoring across evaluations.}
Canonical evaluation and intermediate readouts score each option in its
original position, using the preceding tokens of the unified prompt as
the conditioning prefix. Candidate-state patching instead appends each
complete candidate text to the same prompt and scores that continuation.
The distortion-matched layer interventions measure next-token answer-label
accuracy. We report these scores separately because they evaluate different
predictions: text continuation for candidate preference and label selection
for the layer-intervention accuracy endpoint.

\paragraph{Matching perturbation size.}
For a layer output $H$ and intervened output $\widehat H$, the achieved distortion
is $\|\widehat H-H\|_F$, with mean squared distortion
$\|\widehat H-H\|_F^2/(TD)$. Each random projection or Gaussian control is scaled
along its perturbation direction to match the achieved Frobenius budget of
low-rank truncation for that prompt and layer. The final interpolated control need
not retain rank $r$. The expanded layer intervention measures next-token answer-label
accuracy, whereas candidate-state transfer scores complete candidate texts.
The random orthogonal control uses a Gaussian QR basis
in token space. The token-subspace control instead retains a random subset
of token-coordinate directions, so it is a coordinate-mask projection before
distortion matching. These are distinct perturbation directions.
\par\endgroup
\par\endgroup
\par\begingroup
\Needspace{5\baselineskip}
\subsection{Sampling and Statistical Analysis}
\label{sec:statistical_protocol}
We select validation examples with seed 42 and use the same examples when
comparing prompt formats or interventions. The canonical benchmark uses
120 examples per dataset, except COPA, which provides 100. This gives
1,420 prompts per model and 19,880 prompt--model observations overall.
The larger replication and the intervention experiments use the sample
sizes stated in their respective sections.

For a model with $B$ transformer blocks, relative depth is $d=j/(B-1)$,
where $j=0$ denotes the first block output and $j=B-1$ the final
representation. Embeddings are excluded. The canonical trajectories use
five evenly spaced depth fractions; intermediate readouts and the larger
replication use nine. Fractions are rounded to available layers, whereas
the every-layer controls use all block outputs. Bottleneck agreement is
interpreted at the resolution of the corresponding depth grid.

We report bootstrap 95\% confidence intervals. Paired comparisons resample
the same example pairs together. For pooled candidate-state patching,
we resample models, datasets within models, and candidate pairs within
each setting. The two replacement constructions are treated as separate
pairs when they share a base prompt. Intervention counts include repeated
layer and control evaluations of the same prompts. Capability prediction
instead holds out complete models, as described in
Appendix~\ref{sec:baseline_details}.
\par\endgroup
\FloatBarrier
\Needspace{8\baselineskip}
\section{Geometry across Models and Prompts}
\label{sec:appendix_geometry}
We provide additional maps, prompt controls, and normalization
analyses to test which aspects of the depth profile remain stable.

\FloatBarrier
\Needspace{5\baselineskip}
\subsection{Additional Covariance Visualizations}
\label{sec:appendix_visualizations}

\begin{figure}[htbp]
\centering
\includegraphics[width=\textwidth]{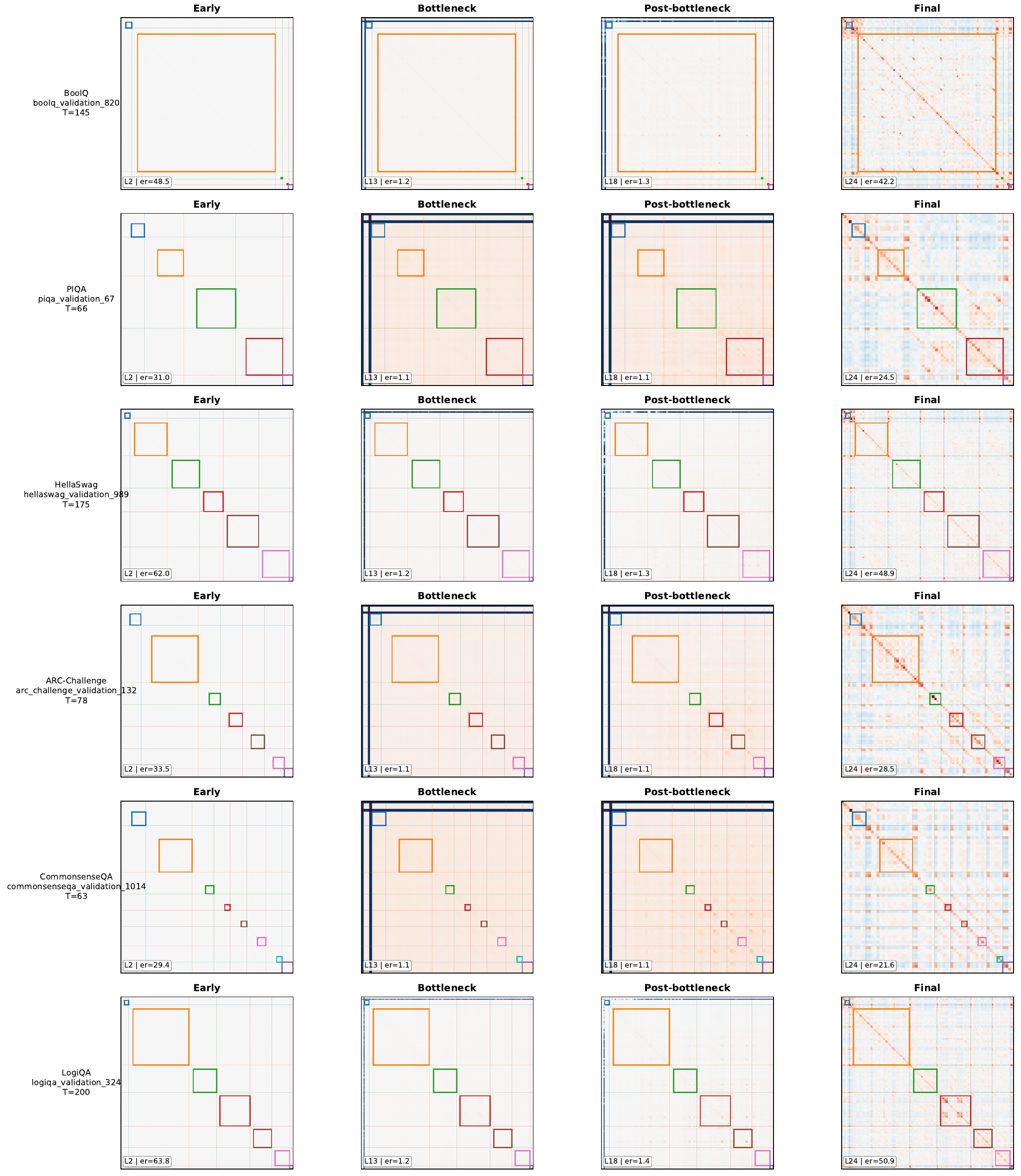}
\caption{\textbf{Representative covariance-map gallery across tasks.} Six deterministic
examples from Pythia-410M show the same broad depth profile beyond the single main-text PIQA example:
early diffuse structure, a sharp bottleneck, and later re-expansion into task-aligned
off-diagonal interactions.}
\label{fig:appendix_gallery}
\end{figure}

\paragraph{Representative covariance maps across tasks.}
We show six deterministic examples from BoolQ, PIQA, HellaSwag,
ARC-Challenge, CommonsenseQA, and LogiQA in Figure~\ref{fig:appendix_gallery}.
Each is shown at an early layer, the bottleneck, a post-bottleneck layer,
and the final layer, allowing comparison with the single PIQA example in
Figure~\ref{fig:heatmap}.

\begin{figure}[htbp]
\centering
\includegraphics[width=\textwidth]{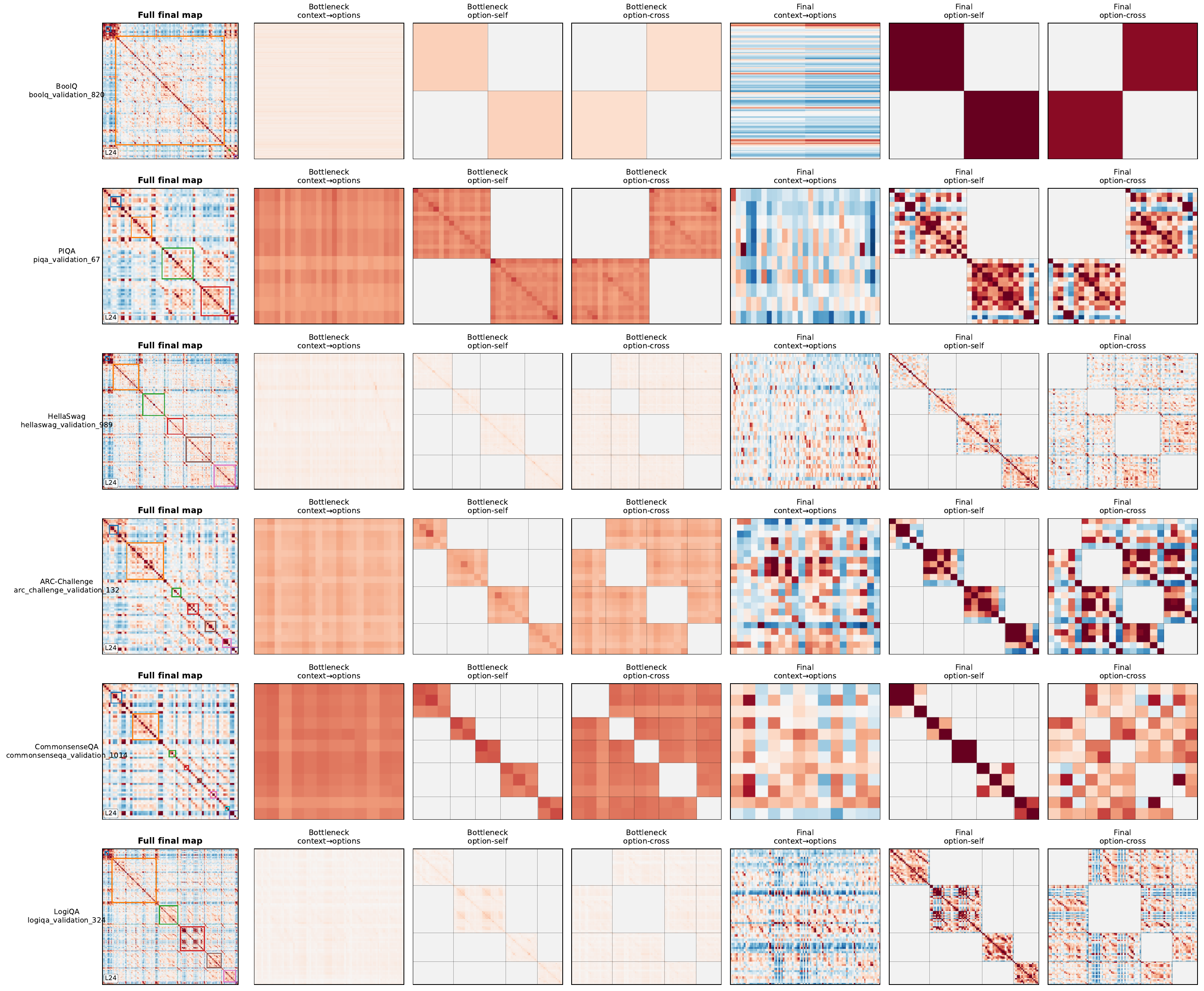}
\caption{\textbf{Segment-level appendix gallery.} For the same representative prompts,
we show the full final-layer map together with context--option,
option-self, and option-cross blocks at the bottleneck and final layer. These zooms make
the shift from compression to comparison visually explicit.}
\label{fig:segment_gallery}
\end{figure}

\paragraph{Segment-level views of context and options.}
Figure~\ref{fig:segment_gallery} isolates the blocks used in our interpretation:
context--option coupling (CtxOpt), within-option structure (OptSelf), and
between-option interaction (OptCross). The same
representative prompts are shown at the bottleneck and final layer.

\begin{figure}[htbp]
\centering
\includegraphics[width=0.85\textwidth,height=0.86\textheight,keepaspectratio]{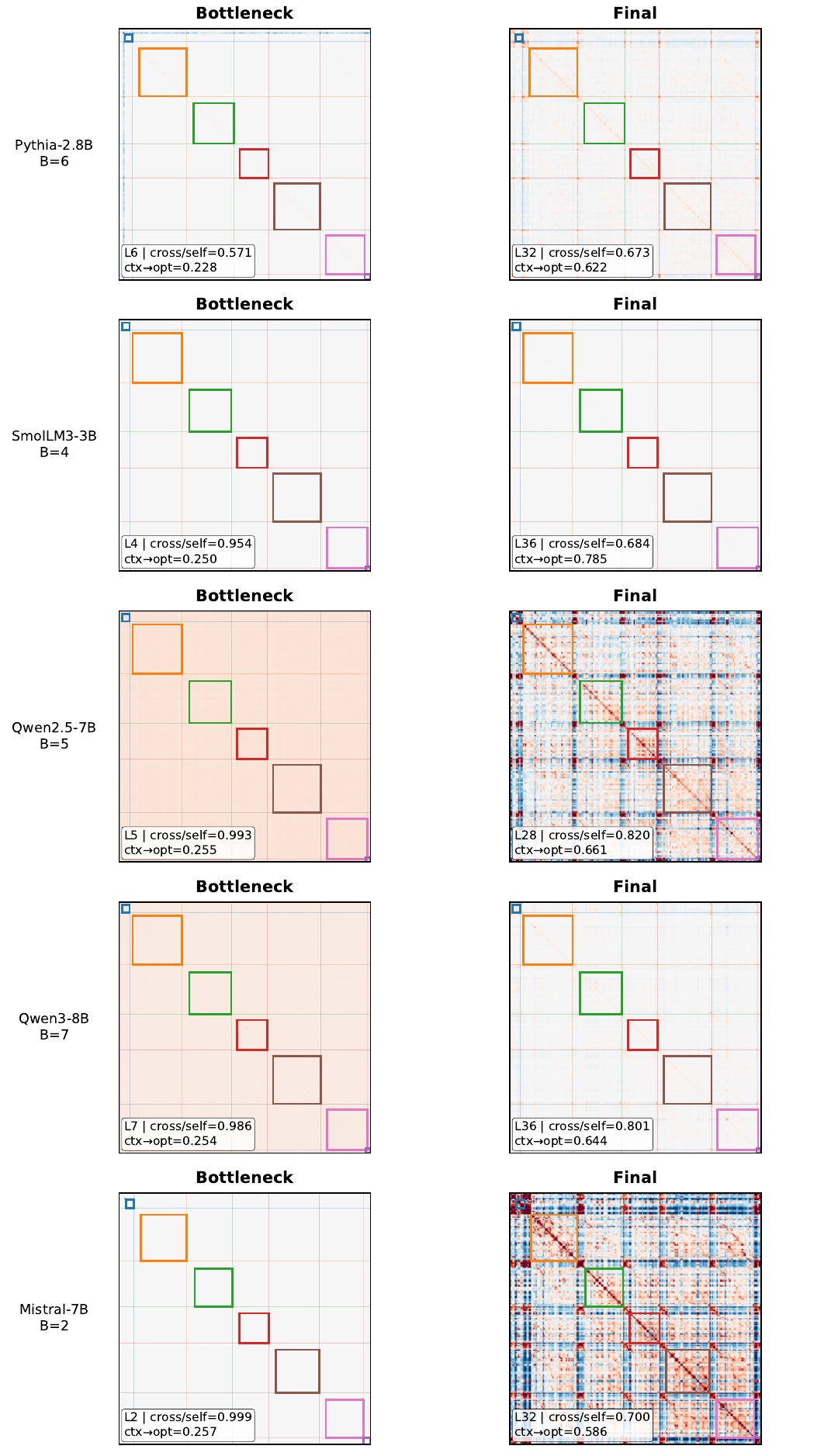}
\caption{\textbf{Open-model comparison on one fixed 4-choice prompt.} Across the
available comparison models, the bottleneck-versus-final-layer contrast remains visually
recognizable even when the exact layer index and cross/self strength vary by model.}
\label{fig:modern_gallery}
\end{figure}

\paragraph{Open-model comparison.}
Figure~\ref{fig:modern_gallery} compares a fixed 4-choice prompt across additional open
models. The result is a qualitative robustness check showing that the same basic
bottleneck-versus-final-layer contrast remains visible outside the original family-scale
benchmark.

\FloatBarrier
\Needspace{5\baselineskip}
\subsection{Robustness of the Compress-then-Compare Profile}
\label{sec:appendix_robustness}

\begin{figure}[htbp]
\centering
\includegraphics[width=\textwidth]{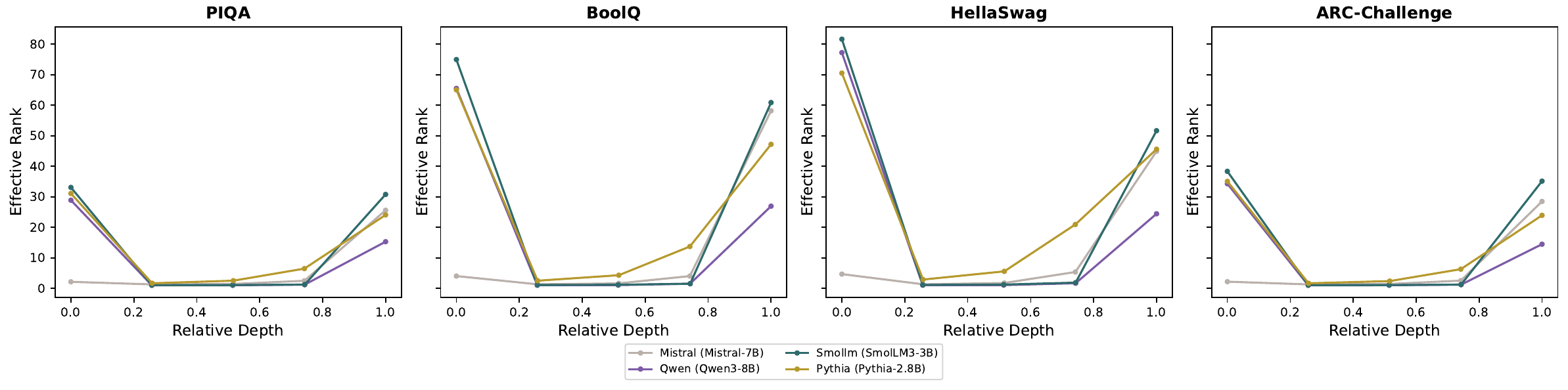}
\caption{\textbf{Cross-family validation.} Using representative models from each family, the same qualitative bottleneck appears across Pythia, SmolLM2, Qwen2.5, and Mistral. This tests whether the depth profile persists across model families, not only Pythia scales.}
\label{fig:cross_family}
\end{figure}

\paragraph{Cross-family validation.}
Figure~\ref{fig:cross_family} compares representative models from each family on
representative tasks. The same effective-rank profile appears across Pythia, SmolLM2,
Qwen2.5, and Mistral, supporting the claim that the depth profile is not tied to a single
model family.

\begin{figure}[htbp]
\centering
\includegraphics[width=\textwidth]{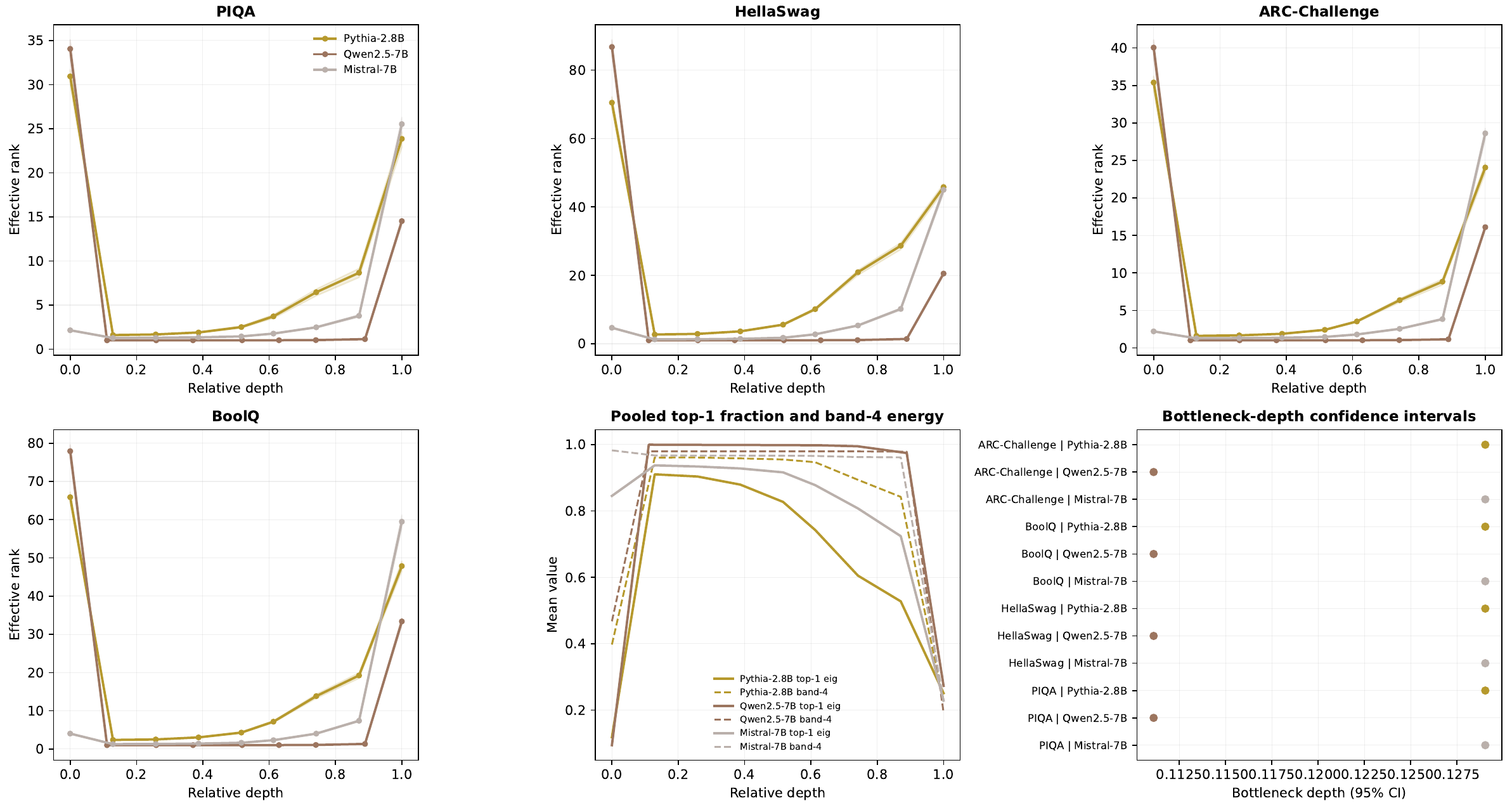}
\caption{\textbf{High-sample fixed-template robustness.} Across three representative models and four datasets, 299--500-example reruns preserve an interior bottleneck and late re-expansion. Under the fixed relative-depth grid, the bootstrap-selected bottleneck depth is stable across 200 resamples per setting.}
\label{fig:high_sample_main_law}
\end{figure}

\paragraph{High-sample robustness of the main law.}
We repeat the fixed-template analysis with Pythia-2.8B, Qwen2.5-7B,
and Mistral-7B-v0.3 on 500 examples each from PIQA, BoolQ, and HellaSwag
and all 299 ARC-Challenge validation examples. We use 200 bootstrap
resamples per setting. All 12 settings retain an interior minimum and
late re-expansion, as shown in Figure~\ref{fig:high_sample_main_law}.
The narrow bottleneck-depth intervals describe stability on the sampled
relative-depth grid; the every-layer analysis below tests finer localization.
Every bottleneck-depth 95\% confidence interval has width
below 0.05 in relative-depth units.

\begin{figure}[htbp]
\centering
\includegraphics[width=\textwidth]{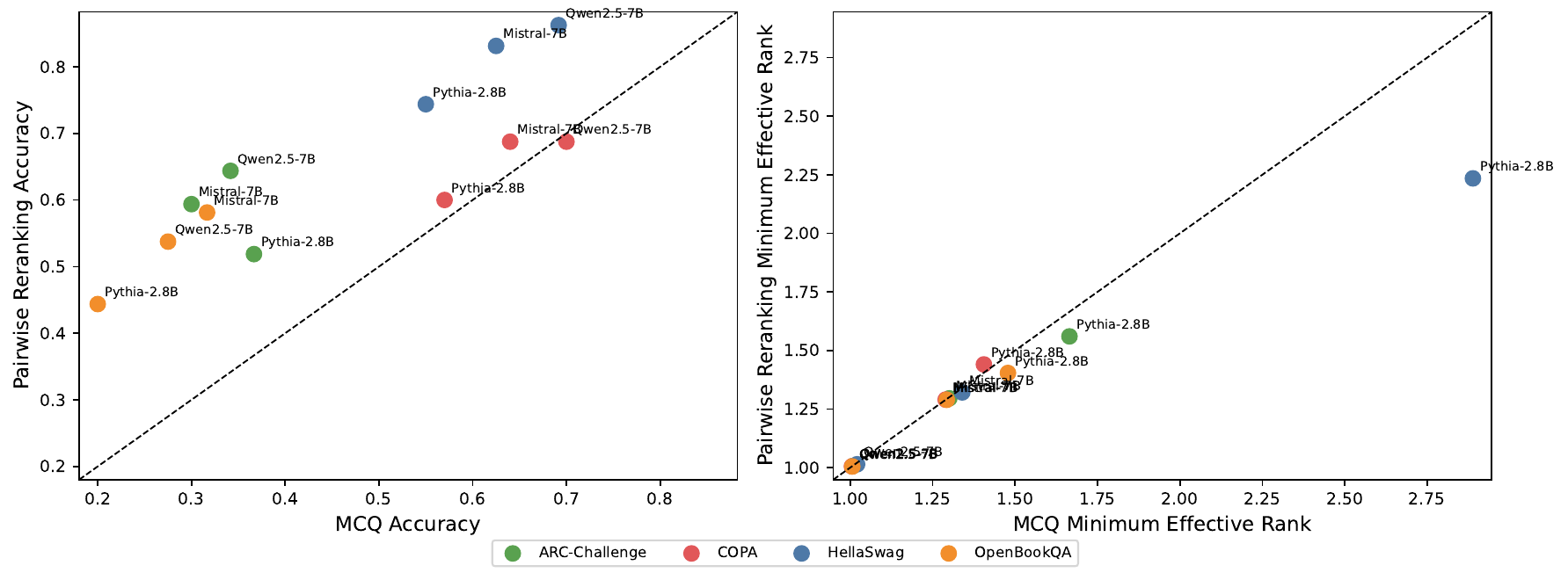}
\caption{\textbf{Pairwise reranking preserves the same bottleneck geometry beyond the canonical multiple-choice template.} Left: reranking often improves behavior on 4-choice tasks. Right: minimum effective rank stays close to the multiple-choice value.}
\label{fig:reranking}
\end{figure}

\paragraph{Pairwise reranking preservation.}
We reformat HellaSwag, COPA, ARC-Challenge, and OpenBookQA into pairwise Candidate A
versus Candidate B prompts. Figure~\ref{fig:reranking} shows that bottleneck depth and
minimum rank are preserved under this reformulation.

\begin{figure}[htbp]
\centering
\includegraphics[width=\textwidth]{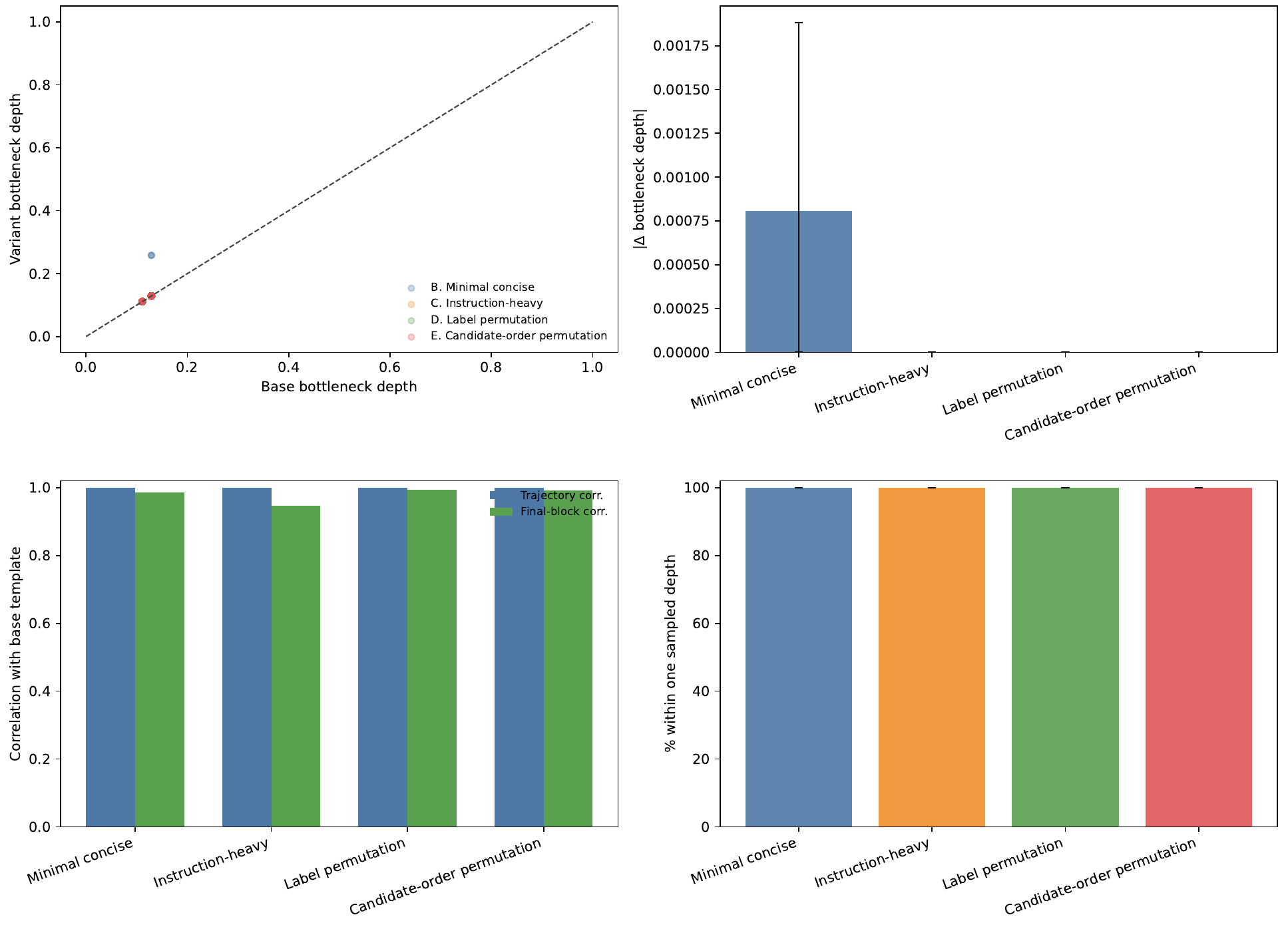}
\caption{\textbf{Deterministic prompt-format robustness.} The variants are deterministic templates, not LLM-generated paraphrases. Across all tested templates, the bottleneck stays within one sampled depth of the base prompt, effective-rank trajectories remain nearly perfectly correlated with the base template, and final-layer block summaries remain highly aligned.}
\label{fig:prompt_robustness}
\end{figure}

\paragraph{Deterministic prompt-format robustness.}
We compare the unified template with concise and instruction-heavy rewrites,
label permutations, and candidate-order permutations. Pythia-2.8B,
Qwen2.5-7B, and Mistral-7B-v0.3 use the same examples from PIQA, BoolQ,
HellaSwag, and ARC-Challenge for each variant. As shown in
Figure~\ref{fig:prompt_robustness}, bottleneck depth and effective-rank
trajectories remain nearly unchanged. Final-layer block summaries are also
closely aligned, although more sensitive to wording than the rank trajectory.

\begin{figure}[htbp]
\centering
\includegraphics[width=\textwidth]{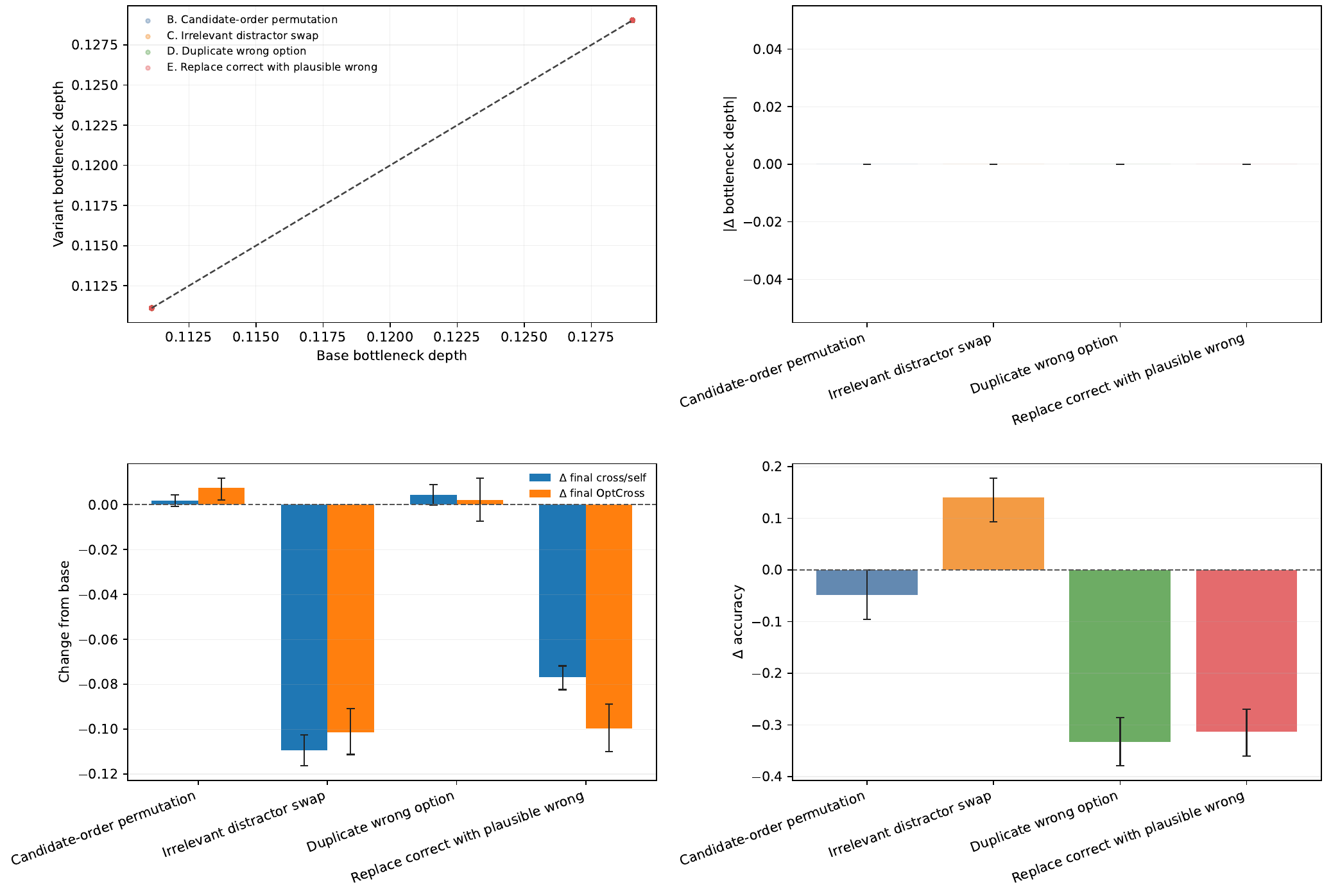}
\caption{\textbf{Controlled answer-set manipulations.} Across all tested variants, the mid-layer bottleneck remains fixed while late comparison geometry becomes sensitive once the candidate set itself changes. Duplicate-wrong is a boundary case: behavior can degrade strongly even when the coarse late-geometry summaries move little.}
\label{fig:answer_set_manipulation}
\end{figure}

\paragraph{Controlled answer-set manipulations.}
We use Pythia-2.8B and Qwen2.5-7B on HellaSwag, ARC-Challenge, and
PIQA to compare candidate reordering, length-matched irrelevant distractors,
duplicate wrong options where valid, and plausible wrong replacements for
the correct option. Figure~\ref{fig:answer_set_manipulation} separates the
two geometric stages: candidate content changes late block summaries,
while the bottleneck remains fixed. Candidate reordering has little effect.
Duplicating a wrong option can instead reduce accuracy without a comparable
change in the pooled block summaries.

For correct-answer replacement, the original gold
position is retained as the reference index even though its text is
replaced. The displayed accuracy is therefore selection of that index,
not semantic correctness on a newly relabeled task. Candidate-order
permutations instead remap the gold index to follow the same answer.

\begin{figure}[htbp]
\centering
\includegraphics[width=\textwidth]{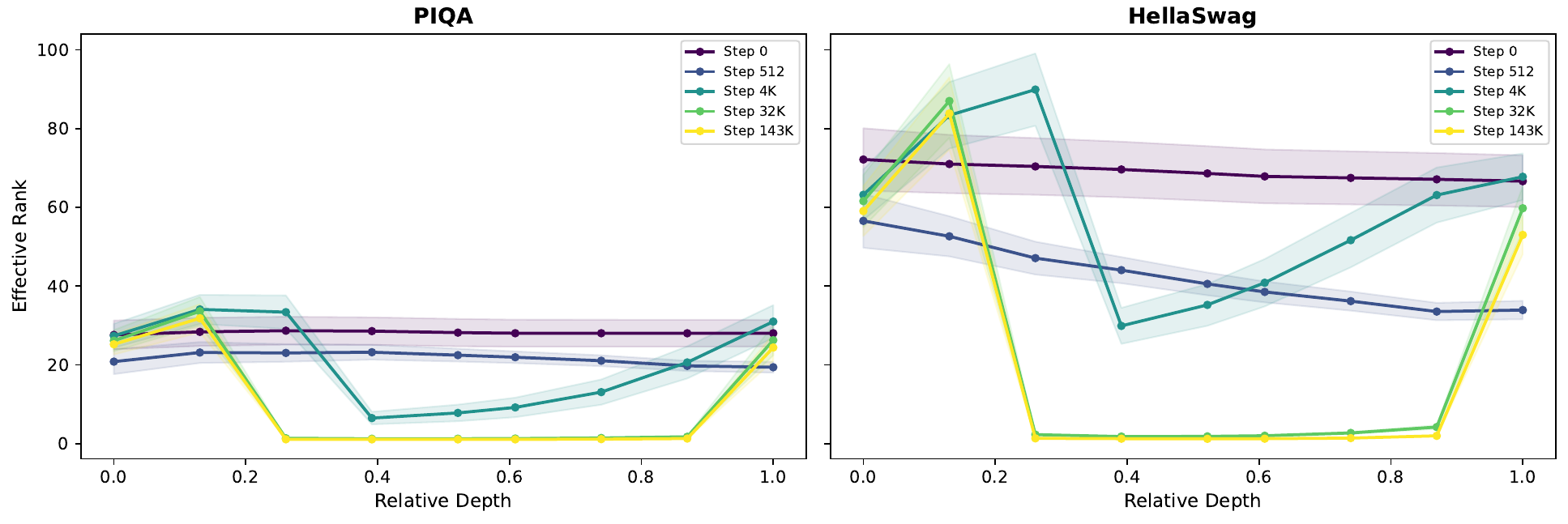}
\caption{\textbf{Training dynamics for Pythia-410M up to 143K steps.} The mid-layer bottleneck is absent at initialization and appears progressively during training, arguing against the simplest explanation that the depth profile is a fixed architectural artifact.}
\label{fig:training}
\end{figure}

\paragraph{Training dynamics.}
Using Pythia checkpoints at steps 0, 512, 4K, 32K, and 143K, we track effective rank
through depth. Figure~\ref{fig:training} shows that the depth profile is nearly flat at
initialization; as training proceeds, the mid-layer bottleneck appears, deepens, and
stabilizes.

\begin{figure}[htbp]
\centering
\includegraphics[width=0.8\textwidth]{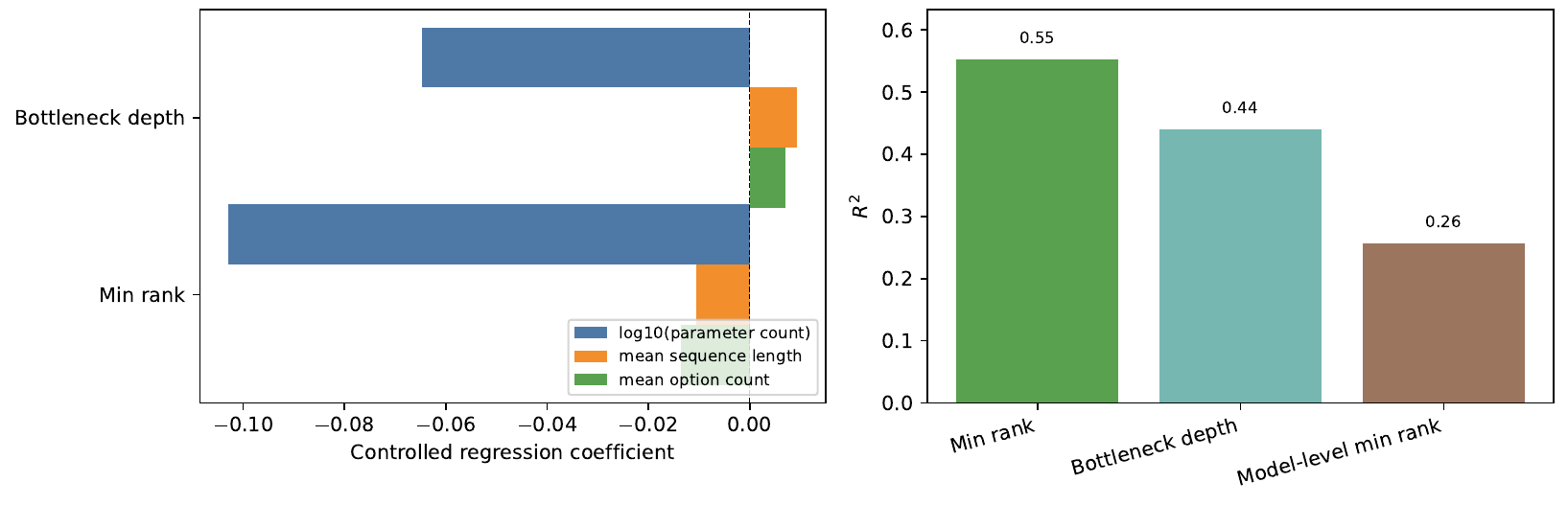}
\caption{\textbf{Length and option-count robustness.} The associations
of compression summaries with model size retain their direction after adjustment
for prompt length and option count.}
\label{fig:length_control}
\end{figure}

\paragraph{Length and option-count controls.}
We regress minimum effective rank and bottleneck depth on model scale while
controlling for mean sequence length and option count. Their relationships
with scale retain the same direction after adjustment, as shown in
Figure~\ref{fig:length_control}. This model-level check complements the
prompt-level normalization analysis in
Section~\ref{sec:normalization}.

\paragraph{Null controls and prompt sensitivity.}
Semantic block effects are stronger than random same-shaped contiguous blocks, especially
in middle and late layers. In addition, the overall phase-transition shape remains
qualitatively stable across multiple prompt templates on PIQA and ARC-Easy.

\begin{figure}[htbp]
\centering
\includegraphics[width=\textwidth]{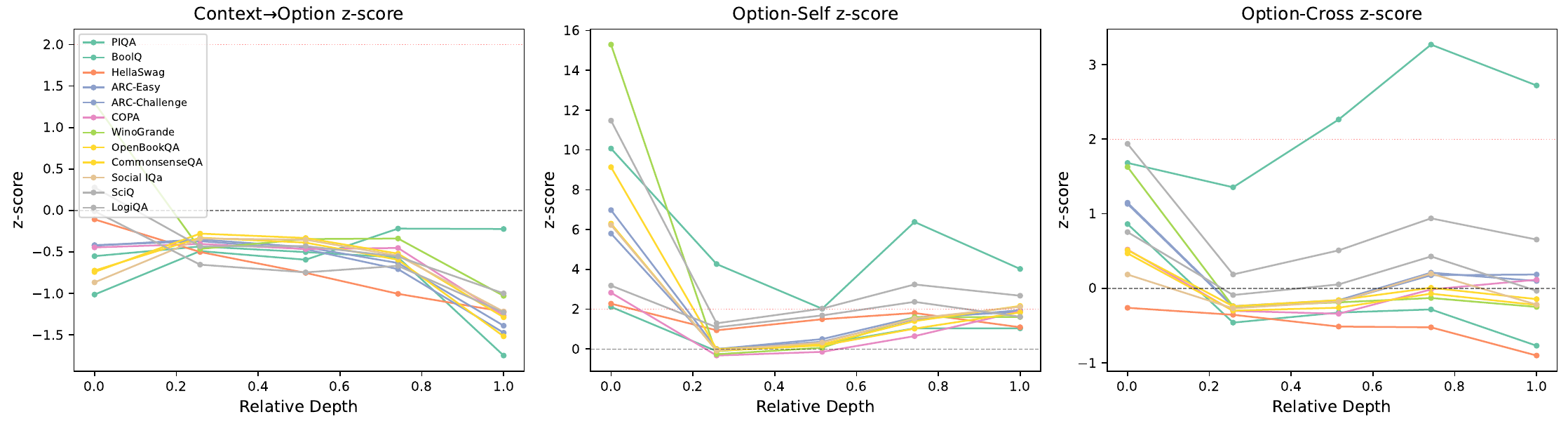}
\caption{\textbf{Spatial null-control z-scores for semantic block structure.} For Pythia-2.8B, semantic block effects are compared with random same-shaped contiguous blocks to check that the block summaries are not purely a shape artifact.}
\label{fig:null}
\end{figure}

Figure~\ref{fig:null} reports the spatial null-control z-scores.

\FloatBarrier
\par\begingroup
\Needspace{5\baselineskip}
\subsection{Every-Layer Prompt and Candidate Controls}
\label{sec:every_layer_controls}
We repeat prompt changes at every transformer layer using the same three
models and four datasets, with up to 120 paired examples per condition.
Concise and instruction-heavy rewrites and candidate reordering are evaluated
on all four datasets. Irrelevant-distractor and correct-answer replacement are
evaluated on PIQA, HellaSwag, and ARC-Challenge; BoolQ does not support these
candidate-content replacements. There are 6,480 paired condition observations.
The minimizing layer is unchanged under all five perturbations, while
compressed-region widths can vary.

\begin{table}[htbp]
\centering
\small
\begin{tabular}{lrrr}
\toprule
Condition & Pairs & Region IoU [95\% CI] & Trajectory correlation\\
\midrule
Concise rewrite & 1,440 & 0.9778 [0.9561,0.9963] & 0.999367\\
Instruction-heavy rewrite & 1,440 & 0.9544 [0.9197,0.9938] & 0.998883\\
Candidate order & 1,440 & 0.9980 [0.9961,0.9996] & 0.999962\\
Irrelevant distractor & 1,080 & 0.9824 [0.9600,0.9979] & 0.998801\\
Correct-answer replacement & 1,080 & 0.9677 [0.9317,0.9954] & 0.997487\\
\bottomrule
\end{tabular}
\caption{\textbf{Every-layer prompt and candidate changes preserve the compressed
region.} These are paired deterministic changes, without LLM-generated
paraphrases. The same examples are used for each valid comparison. The minimizing
layer is unchanged in every condition; region IoU additionally measures changes
in interval width. This analysis is distinct from the sampled-depth template
comparison in \Cref{fig:prompt_robustness}.}
\label{tab:full_layer}
\end{table}

\begin{figure}[htbp]
\centering
\includegraphics[width=\textwidth]{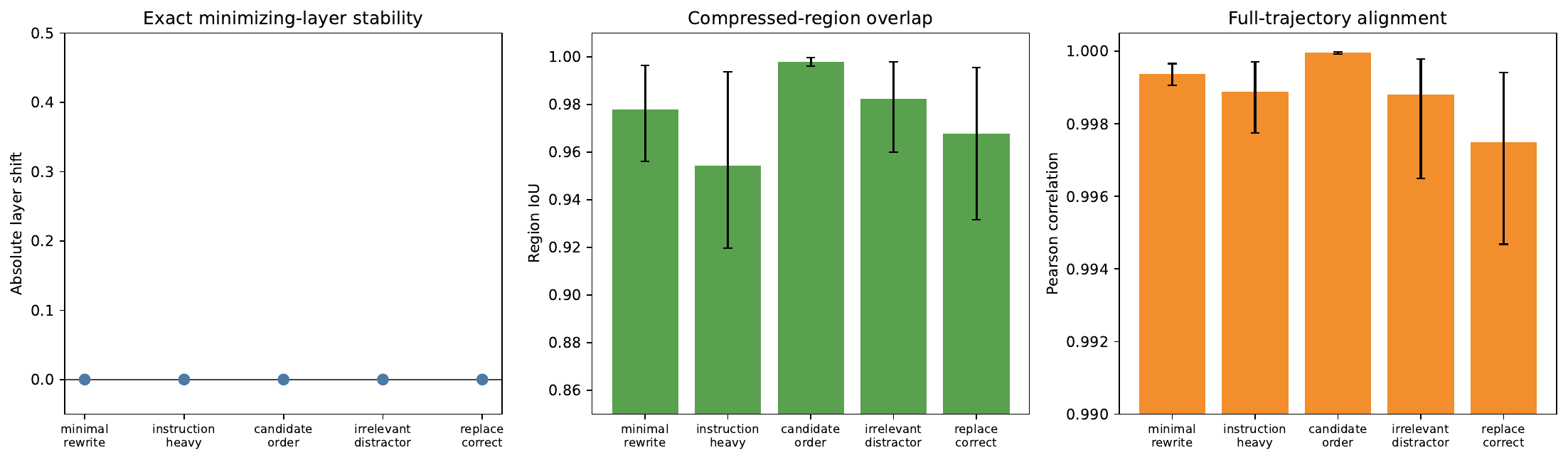}
\caption{\textbf{Region stability at every transformer layer.}
Pythia-2.8B, Qwen2.5-7B, and Mistral-7B-v0.3 on the valid prompt and
candidate perturbations in \Cref{tab:full_layer}. The analysis compares
minimizing-layer shift, compressed-region overlap, and complete rank trajectories.}
\label{fig:full_layer}
\end{figure}
\par\endgroup

\FloatBarrier
\par\begingroup
\Needspace{5\baselineskip}
\subsection{Normalization, Tokenization, and Hidden-State Norms}
\label{sec:normalization}

The normalization and covariate analyses use all 19,880 canonical prompt--model
observations. Covariates are sequence length in tokens, character count, tokens
per character, option count, mean option length, option-length imbalance, and
hidden-state root-mean-square (RMS) norm. Jointly, they explain 5.38\% of
bottleneck-depth variance and 61.98\% of re-expansion-magnitude variance.
The lowest interior-bottleneck rate in any matched covariate bin is 99.879\%.
The existence and location of compression are therefore much more stable than
the raw magnitude of the late rise.

\begin{table}[htbp]
\centering
\small
\begin{tabular}{lrr}
\toprule
Representation & Interior bottleneck & Mean rank re-expansion\\
\midrule
Token covariance & 99.970\% & 23.20\\
Direct hidden-state rank & 99.794\% & 52.57\\
Centered cosine & 91.318\% & 15.91\\
Uncentered Gram & 85.714\% & 7.56\\
\bottomrule
\end{tabular}
\caption{\textbf{Normalization changes magnitude more than the existence of the
transition.} An interior minimum occurs at neither endpoint of the evaluated
trajectory. Rank re-expansion is the increase from minimum to final effective rank within each
representation; its raw units need not coincide across representations.}
\label{tab:normalization}
\end{table}

Centered Gram and covariance differ only by a positive scalar under the stated
centering convention. Positive global scaling leaves normalized eigenvalues,
effective rank, bottleneck depth, and cross/self unchanged. Centered cosine
instead normalizes token norms individually. Absolute block means measure
interaction strength but discard sign; the signed candidate analysis below
retains those distinctions without turning symmetric covariance into a
directional measure.

\begin{figure}[htbp]
\centering
\includegraphics[width=\textwidth]{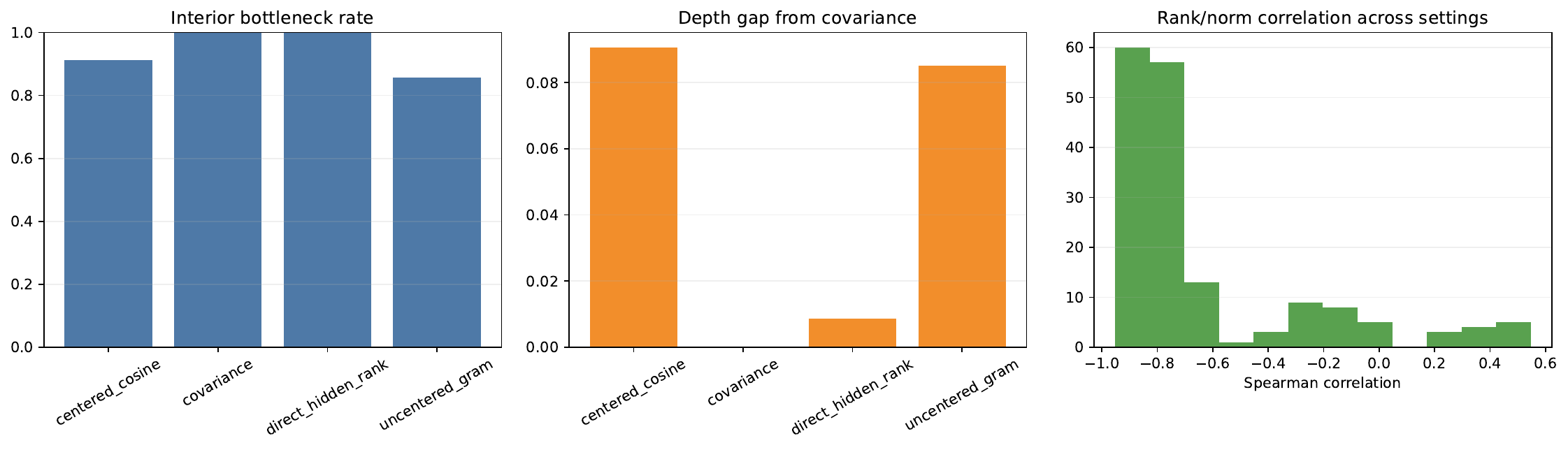}
\caption{\textbf{Normalization and confound analysis.}
The full canonical benchmark is recomputed with alternative spectral and map
summaries. Covariate analysis includes length, tokenization, option statistics,
and hidden-state RMS. Transition location and re-expansion magnitude are
reported separately.}
\label{fig:normalization}
\end{figure}
\par\endgroup
\FloatBarrier
\Needspace{8\baselineskip}
\section{Candidate Content and Late Comparison Geometry}
\label{sec:appendix_content}
These experiments separate changes to instructions and surface
form from changes to candidate content, and retain sign and candidate identity
in the late block analysis.

\par\begingroup
\Needspace{5\baselineskip}
\subsection{Matched Selection Objectives}
\label{sec:objectives}

The objective control isolates the instruction from the rest of the prompt.
Pythia-2.8B, Qwen2.5-7B, and Mistral-7B-v0.3 use the same 200 examples per
PIQA, BoolQ, HellaSwag, and ARC-Challenge setting. Selection and neutral
inspection retain identical context and candidate tokens; only the instruction
changes. Every transformer-block output is measured. The compressed region is
defined in \Cref{sec:methods}.

\begin{table}[htbp]
\centering
\small
\begin{tabular}{lcc}
\toprule
Paired measurement & Estimate & Bootstrap 95\% CI\\
\midrule
Selection/neutral compressed-region IoU & 0.996 & [0.993,1.000]\\
Absolute minimizing-layer shift, layers & 0.000 & [0.000,0.000]\\
Selection vs. neutral: OptCross rise & 0.0484 & [-0.0178,0.1618]\\
Selection vs. neutral: cross/self rise & 0.0070 & [-0.0058,0.0361]\\
W/o candidate content: region IoU & 0.947 & [0.863,0.998]\\
W/o candidate content: OptCross rise & -0.327 & [-0.700,-0.027]\\
W/o candidate content: cross/self rise & -0.073 & [-0.121,-0.006]\\
W/o candidate content: CtxOpt rise & -0.095 & [-0.142,-0.048]\\
\bottomrule
\end{tabular}
\caption{\textbf{Compression is shared across objectives while candidates affect
late structure.} Rise is the change from the bottleneck to the final layer.
W/o candidate content replaces candidates with approximately length-matched
placeholders while retaining the context; its changes are relative to the
selection prompt. For selection vs. neutral, positive values favor selection.
IoU measures overlap of compressed layer
intervals. Intervals use 10,000 hierarchical bootstrap replicates.}
\label{tab:objectives}
\end{table}

Selection and neutral inspection have nearly identical compressed regions.
Their late-geometry differences have intervals crossing zero. Removing candidate
content with approximately length-matched placeholders retains compression while
reducing the late block rises in \Cref{tab:objectives}. The objective
comparison identifies shared compression, rather than a universal late increase
caused by a selection verb.

We distinguish the shared compressed region from the
late comparison structure. Neutral inspection retains compression, while
candidate-content controls, signed blocks, and candidate-state patching
connect the later structure to the candidate set and subsequent preference,
as shown in \Cref{tab:objectives,tab:signed,fig:candidate_patching}.

\begin{figure}[htbp]
\centering
\includegraphics[width=\textwidth]{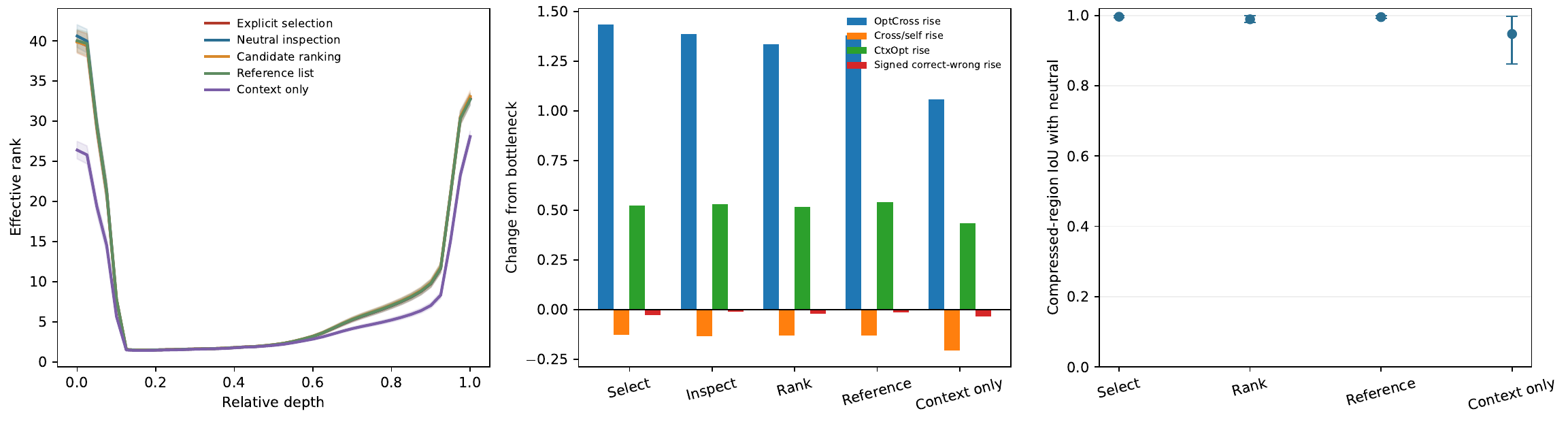}
\caption{\textbf{Objective and candidate-content controls.}
Panels show effective-rank trajectories, bottleneck-to-final block changes, and
compressed-region IoU (left to right) across selection, inspection, ranking,
reference-list, and candidate-content conditions.
Every non-instruction token is fixed for selection versus neutral inspection.
The context-only placeholder condition uses approximate length matching.
The supported distinction is shared compression and candidate-sensitive late
structure.}
\label{fig:objectives}
\end{figure}
\par\endgroup

\FloatBarrier
\par\begingroup
\Needspace{5\baselineskip}
\subsection{Signed Candidate Structure}
\label{sec:semantics}

The candidate-specific analysis retains sign and separates correct and wrong
options. In place of the absolute mean $S_\ell(A,B)$, use the signed mean
$|A|^{-1}|B|^{-1}\sum_{i\in A,j\in B}C_{ij}^{(\ell)}$ and average separately
over context-correct, context-wrong, correct-self, wrong-self, correct-wrong,
and wrong-wrong blocks. Across all 19,880 observations, the three
bottleneck-to-final contrasts in \Cref{tab:signed} exclude zero.

\begin{table}[htbp]
\centering
\small
\begin{tabular}{lrr}
\toprule
Signed contrast, bottleneck-to-final change & Mean change & Bootstrap 95\% CI\\
\midrule
Context-correct vs. context-wrong & -0.0450 & [-0.0776,-0.0108]\\
Correct-self vs. wrong-self & -0.0594 & [-0.1205,-0.0027]\\
Correct-wrong vs. wrong-wrong & -0.0353 & [-0.0710,-0.0015]\\
\bottomrule
\end{tabular}
\caption{\textbf{Correct and wrong candidate relations evolve differently after
compression.} Each contrast is the signed covariance difference between
the first and second named blocks. Signs measure covariance association, not direct evidence for a
candidate or directed information flow. The interventions in
\Cref{sec:patching_details} test later behavioral influence.}
\label{tab:signed}
\end{table}

\begin{figure}[htbp]
\centering
\includegraphics[width=\textwidth]{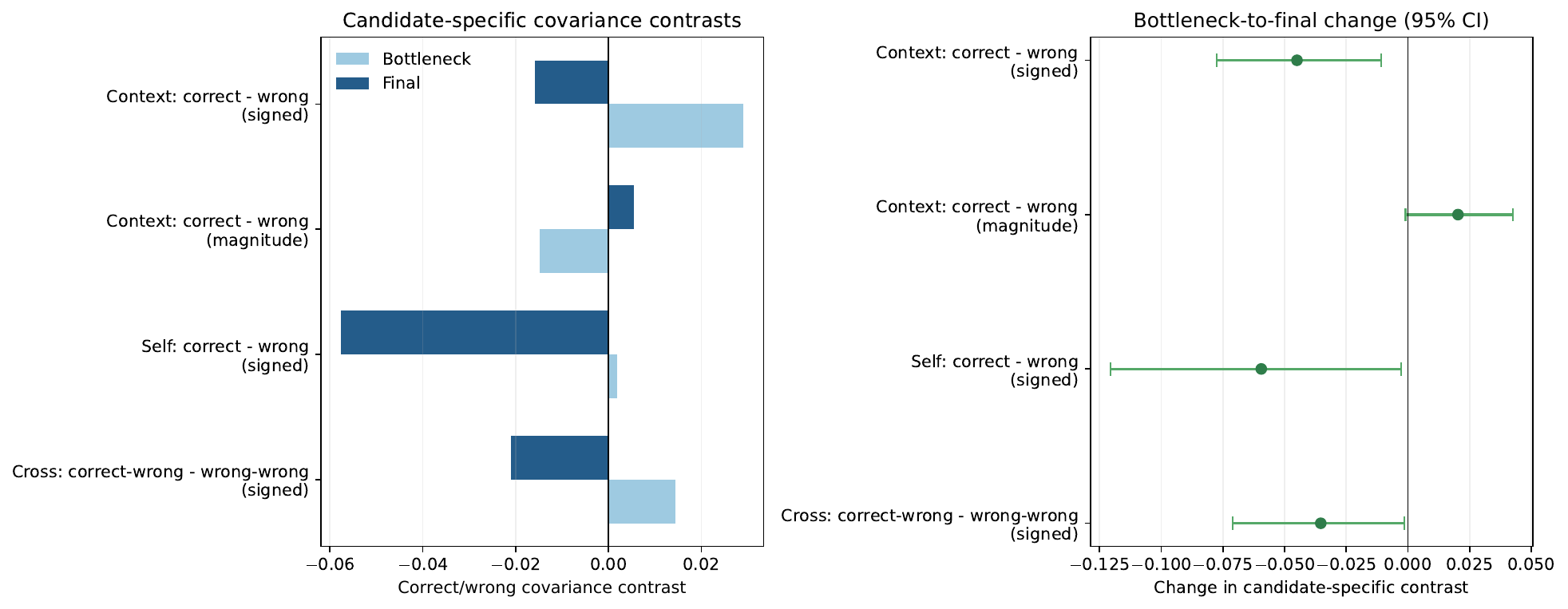}
\caption{\textbf{Signed candidate-specific geometry through depth.}
Context and candidate blocks are separated by correctness, retaining relations
that pooled absolute block summaries average together.}
\label{fig:signed}
\end{figure}

\Needspace{5\baselineskip}
\subsection{Position-Matched Surface Controls}
Pythia-2.8B, Qwen2.5-7B, and Mistral-7B-v0.3 on PIQA, HellaSwag,
ARC-Challenge, and OpenBookQA give 720 model--example pairs at 60 examples per
setting. Within-segment option shuffles preserve the token multiset and segment boundaries;
shape-matched nonsense preserves character and word shape rather than exact
tokenizer length. These controls accompany the random same-shaped contiguous
block comparisons in \Cref{fig:null}. Shuffling and nonsense also disrupt
fluency and local syntax, so they are surface controls rather than clean
isolation of meaning.

\Needspace{5\baselineskip}
\subsection{Exact-Token Semantic Pairs}
We compare a meaning-changing relation reversal with a meaning-preserving entity
rename. Valid pairs have exactly the same total and per-candidate token counts.
There are 30 valid Pythia-2.8B pairs and 20 valid Qwen2.5-7B pairs.
Relation reversal changes late signed candidate geometry more than entity
renaming in both models, as shown in Figure~\ref{fig:exact_semantics},
without shifting the bottleneck. The comparison concerns relational content
under exact token-count matching; the natural-language perturbations in
Section~\ref{sec:natural_language_perturbations} remain mixed.
\par\begingroup
\paragraph{A token-matched relational example.}
One retained Pythia-2.8B triplet uses the context ``Alice is left of Bob.
Who is on the left?'' with candidates Alice and Bob. Relation reversal changes
``left of'' to ``right of'' while retaining the two candidates and their
positions. The surface control replaces Alice and Bob consistently with Clara
and David in both context and candidates. The correct candidate changes from Alice to Bob
only for relation reversal. Across all retained triplets, total and per-candidate
token counts are matched separately for each model; eligibility under one
tokenizer is assessed independently of eligibility under the other.

The signed statistic compares context--correct and context--wrong covariance,
normalized by the mean absolute covariance of the map. Its correct/wrong
identities follow the gold answer in each condition. The relation-reversal
effect therefore measures a label-aware summary, which includes the change
in the identity of the correct candidate.
\par\endgroup

\begin{figure}[htbp]
\centering
\includegraphics[width=0.9\textwidth]{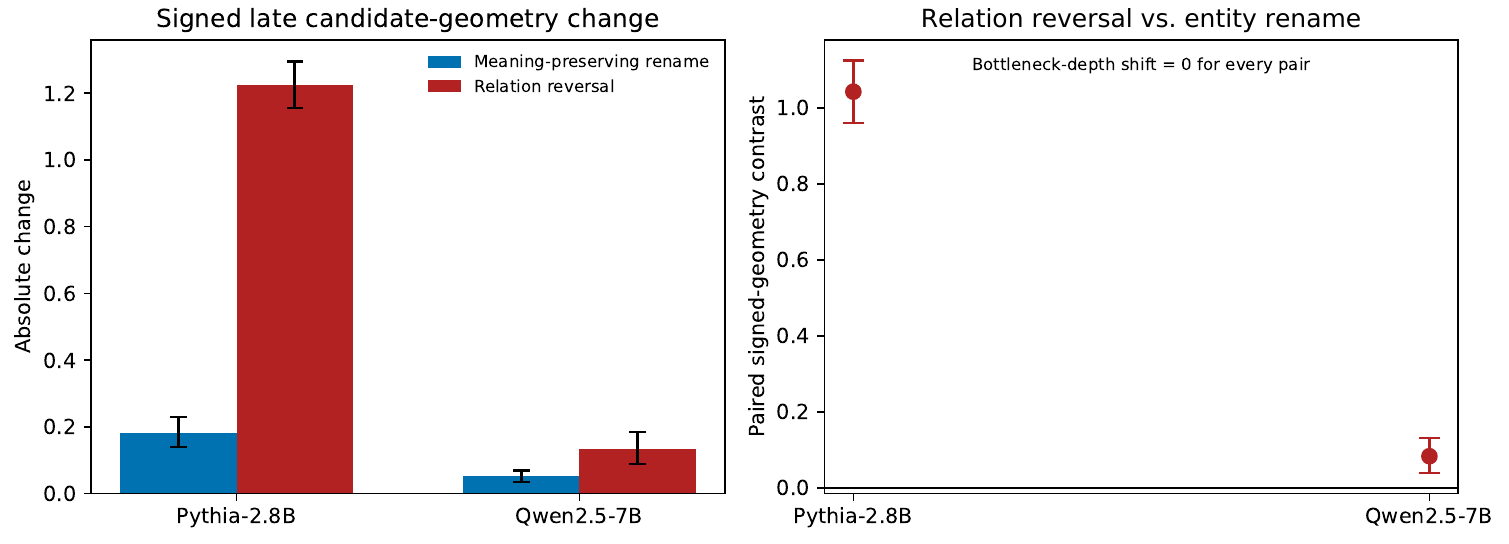}
\caption{\textbf{Relational changes affect late signed geometry under exact token
matching.} Relation reversal is compared with entity renaming on 30 valid
Pythia-2.8B and 20 Qwen2.5-7B pairs. Total and per-candidate token counts are
identical within the paired comparison. The bottleneck remains fixed.}
\label{fig:exact_semantics}
\end{figure}

The duplicate-wrong example in \Cref{fig:answer_set_manipulation} remains an
explicit limit: accuracy can fall markedly while pooled cross/self changes little. Candidate identity and signed views add resolution, but neither
they nor the pooled summaries completely determine behavioral outcomes.
\par\endgroup

\FloatBarrier
\Needspace{5\baselineskip}
\subsection{Natural-Language Perturbations}
\label{sec:natural_language_perturbations}
For each example, we compare meaning-preserving surface rewrites, meaning-changing edits,
and matched superficial controls. Meaning-changing perturbations reduce late-layer
cross/self relative to the superficial control, but meaning-preserving rewrites also move
the geometry strongly and sometimes more than meaning-changing edits. These perturbations mix content changes with surface changes, so they do not
isolate semantic effects.

\begin{figure}[htbp]
\centering
\includegraphics[width=\textwidth]{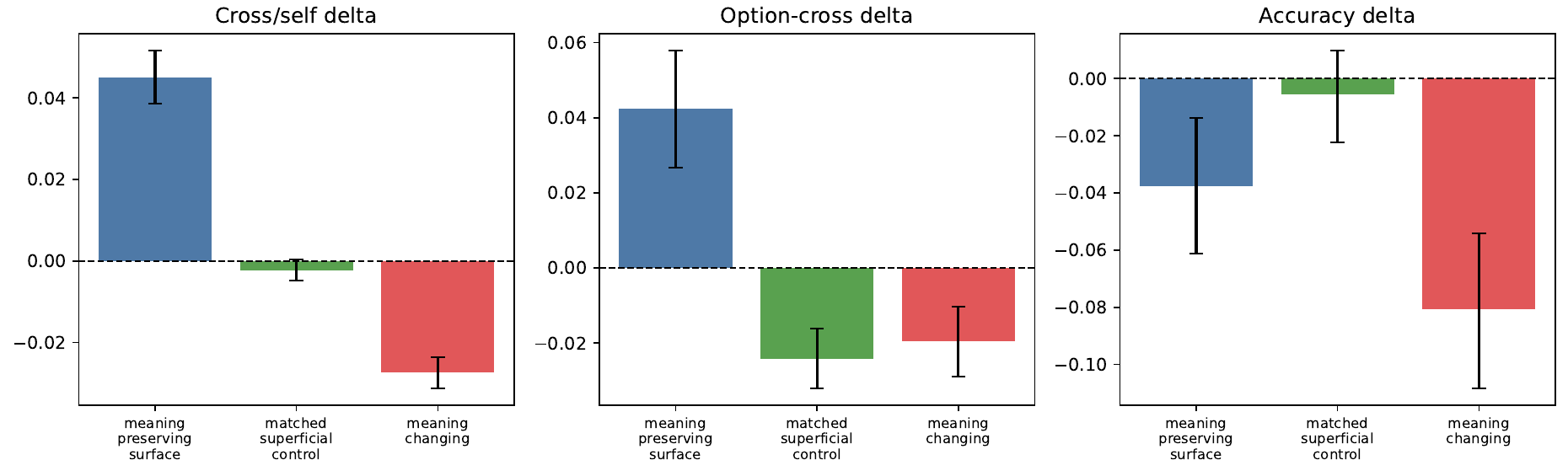}
\caption{\textbf{Matched semantic perturbation analyses.} Meaning-changing edits alter late comparison geometry more than the superficial control, but meaning-preserving surface rewrites also induce large geometry shifts, so the comparison does not isolate semantic content from surface form.}
\label{fig:counterfactuals}
\end{figure}

Figure~\ref{fig:counterfactuals} shows the matched perturbation summary.

\FloatBarrier
\Needspace{8\baselineskip}
\section{Behavioral and Mechanistic Analyses}
\label{sec:appendix_mechanistic}
We distinguish the timing of answer discrimination, distortion-matched
layer interventions, and candidate-specific effects on subsequent preference.

\Needspace{5\baselineskip}
\subsection{Behavioral Timing}
We evaluate Pythia-2.8B, Qwen2.5-7B, and Mistral-7B-v0.3
at nine relative depths. In each model, peak intermediate accuracy occurs
after the covariance bottleneck, as shown in
Figure~\ref{fig:functional_timing}.

\Needspace{5\baselineskip}
\subsection{Layer-Wide Interventions}
Among the original layer-wide interventions, low-rank truncation
gives a more consistent geometric effect than state replacement or layer skipping.
The candidate-specific intervention in \Cref{sec:patching_details} tests
a different endpoint: whether changing the states of one candidate changes subsequent
preference for its complete text.

\begin{figure}[htbp]
\centering
\includegraphics[width=\textwidth]{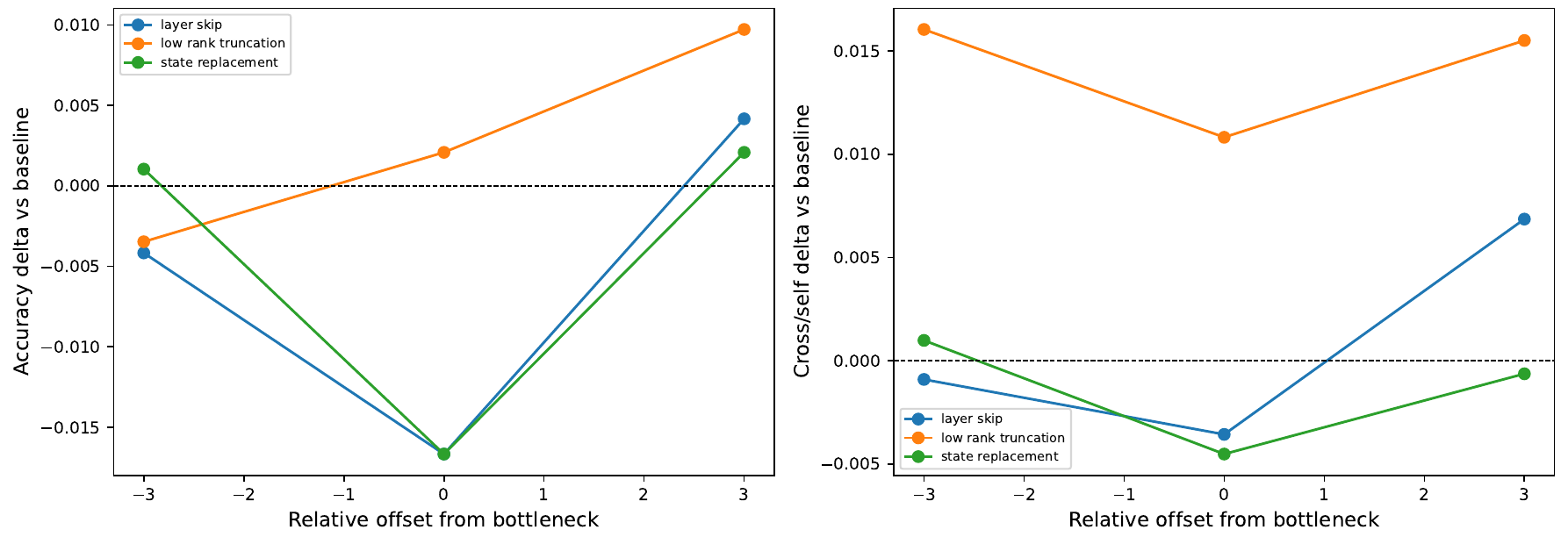}
\caption{\textbf{Original layer-wide interventions around the empirical
bottleneck.} Left: accuracy changes; right: final cross/self changes, each
relative to the unmodified pass. Low-rank truncation yields a more consistent geometric damage pattern
than state replacement or layer skipping. Distortion-matched controls and
candidate-specific behavioral interventions are reported separately in
\Cref{fig:distortion_mech,fig:candidate_patching}.}
\label{fig:mech_appendix}
\end{figure}

Figure~\ref{fig:mech_appendix} compares these intervention families.

\Needspace{5\baselineskip}
\subsection{Dense Layer Sweep}
\begin{figure}[htbp]
\centering
\includegraphics[width=\textwidth]{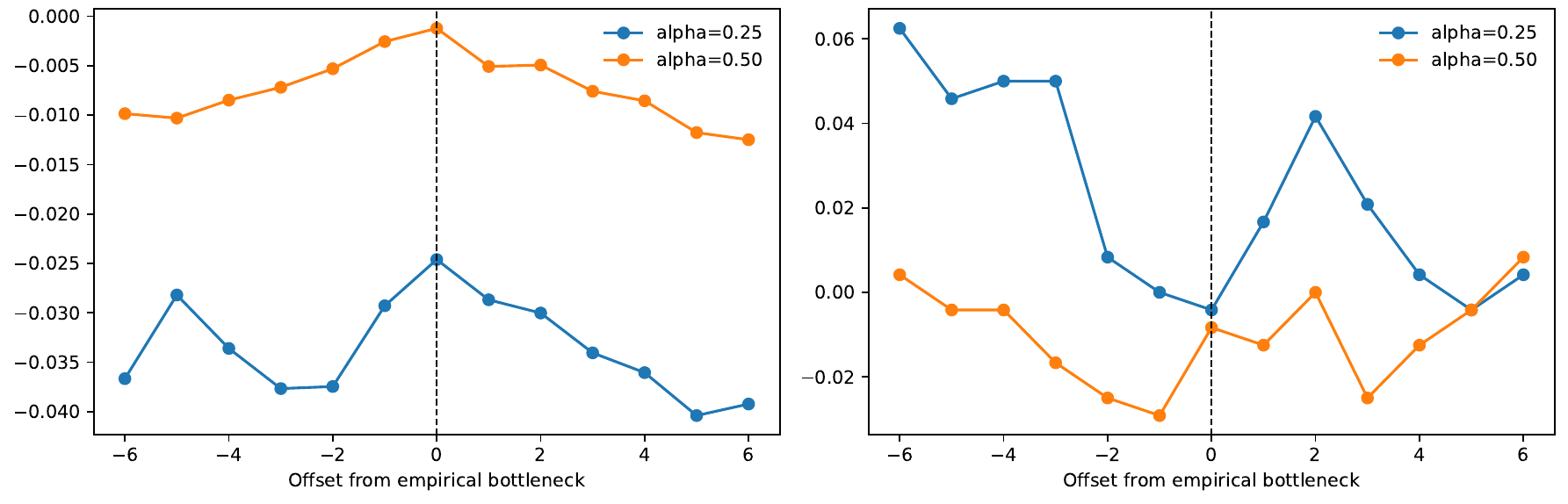}
\caption{\textbf{Dense low-rank truncation sweep around the empirical bottleneck.} Left: final cross/self damage; right: accuracy damage. Damage extends across the bottleneck region, with the largest average effects often occurring at nearby later layers. The sweep therefore locates functional contributions across a region.}
\label{fig:dense_mech}
\end{figure}

We sweep low-rank truncation across offsets from $-6$ to $+6$ for
Pythia-2.8B and Qwen2.5-7B on PIQA, HellaSwag, and ARC-Challenge.
Figure~\ref{fig:dense_mech} shows disruption across the bottleneck region,
with larger effects often occurring at later offsets. Only one of the
12 model--dataset--rank settings has its largest mean cross/self damage
within one layer of the empirical bottleneck. We therefore interpret the
effect as a contribution of the compressed region to later geometry.

\Needspace{5\baselineskip}
\subsection{Distortion-Matched Mechanistic Controls}
\par\begingroup
The expanded benchmark uses Pythia-2.8B and Qwen2.5-7B on PIQA, HellaSwag,
and ARC-Challenge, with 200 prompts per model--dataset pair. We intervene at
offsets $-6,-3,0,+3,+6$ from the empirical bottleneck, retain 25\% or 50\%
of token-space dimensions, and compare empirical top-rank truncation with random
orthogonal, random token-subspace, and isotropic Gaussian perturbations.
Matching the achieved Frobenius norm for every prompt and layer gives 48,000
intervention evaluations. The matching scales a perturbation direction, so a matched
random projection is not necessarily itself a rank-$r$ output after interpolation.
Achieved-to-target norm ratios lie between 0.99990 and 1.00008.

At 25\% retained rank, bottleneck truncation reduces final cross/self by about
0.12 and answer-label accuracy by 7.17 percentage points, as shown in
Figure~\ref{fig:distortion_mech}. To compare regions, we average offsets $-3,0,+3$ and contrast
them with offsets $-6,+6$ on the same prompts.
\Cref{tab:distortion} distinguishes this paired regional contrast from
the regression adjusted for achieved perturbation size, model, and dataset.

\begin{table}[htbp]
\centering
\small
\begin{tabular}{lrr}
\toprule
Bottleneck truncation, retain 25\% & Mean damage & Bootstrap 95\% CI\\
\midrule
Final cross/self & 0.1197 & [0.1122,0.1271]\\
Answer-label accuracy, percentage points & 7.17 & [3.92,10.50]\\
\bottomrule
\end{tabular}
\caption{\textbf{Geometry and behavior under bottleneck truncation.}
Damage is the reduction from the unmodified value after intervention, pooled over
Pythia-2.8B and Qwen2.5-7B on PIQA, HellaSwag, and ARC-Challenge.
Each of the six model--dataset settings uses 200 prompts.
The accuracy endpoint uses answer-label tokens, not the full-text scores of
candidate-state patching.}
\label{tab:distortion_bottleneck_endpoints}
\end{table}

\begin{table}[htbp]
\centering
\small
\begin{tabular}{lcc}
\toprule
Cross/self damage comparison & Retain 25\% & Retain 50\%\\
\midrule
Near vs. distant layers & $0.0584\ [0.0518,0.0648]$ & $0.0549\ [0.0481,0.0614]$\\
Adjusted near-region coefficient & $0.0468\ [0.0408,0.0528]$ & $0.0336\ [0.0269,0.0406]$\\
Low rank vs. random orthogonal & $0.0774\ [0.0729,0.0821]$ & $0.0866\ [0.0825,0.0908]$\\
Low rank vs. token-subspace & $0.0908\ [0.0861,0.0955]$ & $0.1132\ [0.1091,0.1171]$\\
Low rank vs. Gaussian & $-0.0184\ [-0.0248,-0.0115]$ & $0.0010\ [-0.0047,0.0064]$\\
\bottomrule
\end{tabular}
\caption{\textbf{Regional geometry damage and matched controls.}
Damage is the reduction in final cross/self after intervention. Positive
low-rank--control differences mean greater geometric damage from low-rank
truncation. Control contrasts average the near-region offsets.
Bracketed values are bootstrap 95\% confidence intervals. The adjusted coefficient
uses log distortion, model, and dataset covariates and 1,000 prompt-cluster
bootstrap replicates. Low rank does not exceed every equal-distortion control.}
\label{tab:distortion}
\end{table}

The accuracy effect likewise does not consistently exceed every matched control.
Only 2/12 model--dataset--rank settings beat all controls on the behavioral
criterion. The low-rank experiment identifies a region involved in constructing
late comparison geometry; candidate-state transfer provides the direct
behavioral result. Neither experiment identifies uniquely important covariance
directions or a uniquely maximally damaging layer.

\begin{figure}[htbp]
\centering
\includegraphics[width=\textwidth]{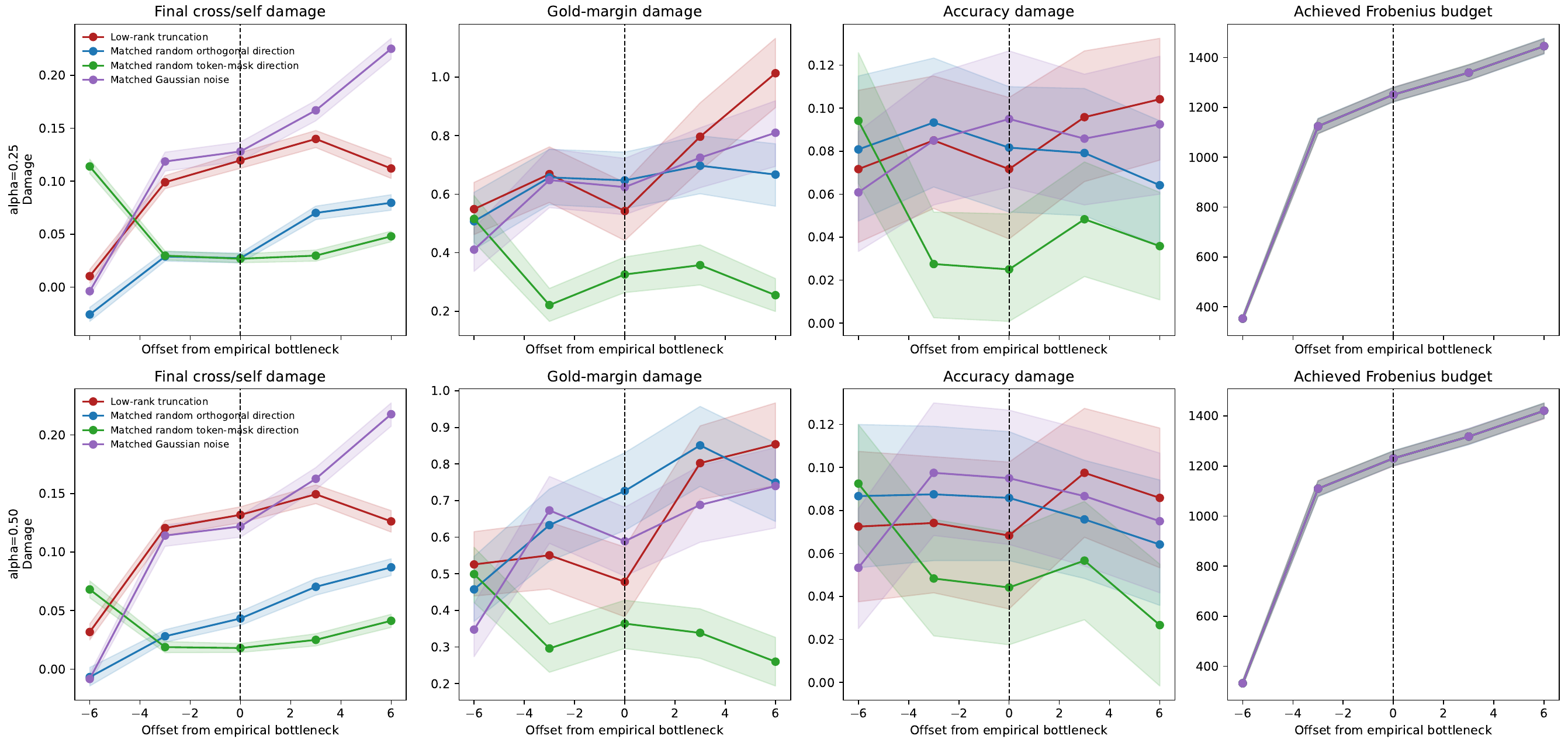}
\caption{\textbf{Distortion-matched layer interventions.}
Final cross/self damage, gold-margin damage, accuracy damage, and
achieved distortion are compared across layer offsets for empirical low-rank truncation, random orthogonal and token-space
projections, and Gaussian perturbations. Gold margin is the log-probability
difference between the correct answer label and the strongest incorrect label.
The expanded analysis uses two models,
three datasets, two rank fractions, and matched per-prompt perturbation budgets.
The conclusion concerns regional construction of late geometry; accuracy effects
do not consistently exceed every control.}
\label{fig:distortion_mech}
\end{figure}
\par\endgroup
\FloatBarrier
\par\begingroup
\Needspace{5\baselineskip}
\subsection{Candidate-State Patching}
\label{sec:patching_details}

The behavioral experiment changes one candidate representation while preserving
the remainder of the prompt. A replacement is drawn from the same dataset,
matched exactly in tokenizer length at the candidate position, and inserted into
a paired prompt. We transfer only the residual states of that span back to the original
prompt. The candidate set scored after transfer contains all original texts plus
the replacement. The primary endpoint is the change in mean continuation log
probability of the replacement relative to the highest-scoring original candidate.

Pythia-2.8B and Qwen2.5-7B use PIQA, HellaSwag, and ARC-Challenge, with up
to 200 base examples per dataset, two replacement constructions, and layer offsets
$-6,-3,0,+3,+6$. Correct-to-plausible-wrong changes a correct candidate to a
wrong candidate; wrong-to-strong-distractor changes an incorrect candidate to a
plausible competing text. A deterministic 20\% development split fixes all choices.
Only the untouched 80\% evaluation split contributes to the 23,605 intervention
evaluations reported here.

\begin{table}[htbp]
\centering
\small
\begin{tabular}{lrr}
\toprule
Candidate-state patching vs. control & Margin difference & Bootstrap 95\% CI\\
\midrule
Unrelated candidate & 3.481 & [2.391,4.651]\\
Same-label unrelated candidate & 3.475 & [2.356,4.642]\\
Matched context position & 0.493 & [0.185,0.957]\\
Alternative position & 0.636 & [0.226,1.294]\\
Norm-matched Gaussian & 3.666 & [2.467,4.918]\\
Randomized candidate state & 3.614 & [2.436,4.859]\\
\bottomrule
\end{tabular}
\caption{\textbf{Candidate content affects full-text preference beyond matched
controls.} Positive values mean a larger donor-margin increase from the changed
candidate patch. The donor is the replacement candidate. Each estimate uses
the paired evaluation observations available for that control at the bottleneck.}
\label{tab:patching_controls}
\end{table}

The unrelated control uses a different candidate; the same-label control selects
a source example with the same gold-answer label. Position controls patch
equally sized context or alternative-position spans. Gaussian and randomized-state
controls match the perturbation norm. Matched-context comparisons use the paired
subset with sufficiently long context spans; other controls retain all otherwise
valid pairs. Identity transfer gives zero maximum logit difference.
Achieved-to-target Frobenius ratios range from 0.9999746 to 1.0000036.
The alternative-position control uses an equally sized span
from the instruction and context, excluding candidate positions, with a
fixed starting position in each prompt.

The donor-margin difference between candidate-state patching and its control
has a positive 95\% confidence
interval in 70 of the 72 prespecified comparisons. These comparisons cross
two models, three datasets, two replacement constructions, and six non-identity
controls at the bottleneck. Table~\ref{tab:patching_endpoints}
reports the pooled behavioral effects and the layer comparisons. Preference
changes extend across the compressed-to-post-compressed region rather than
peaking uniquely at the minimum-rank layer. Multiple offsets and controls are
repeated interventions on the same examples, not independent prompts.

\begin{table}[htbp]
\centering
\small
\begin{tabular}{lrr}
\toprule
Held-out measurement & Estimate & Bootstrap 95\% CI\\
\midrule
Donor-margin change at bottleneck & 3.723 & [2.496,5.041]\\
Directed prediction switches at bottleneck & 9.87\% & [3.29\%,18.09\%]\\
\bottomrule
\end{tabular}
\caption{\textbf{Held-out candidate preference after candidate-state patching.}
The donor is the replacement candidate, and its margin is measured in
mean full-text log-probability units. A directed switch is a change to
the donor as top prediction. Layerwise effects are shown in
Figure~\ref{fig:candidate_patching}, with effects extending across the compressed region.}
\label{tab:patching_endpoints}
\end{table}

\FloatBarrier

\Needspace{5\baselineskip}
\subsection{Residual-Sublayer Localization}
\label{sec:residual_sublayers}
On the same two models and three datasets, we record the residual stream before
attention, after attention, and after the MLP at every layer. The normalized
analysis uses 240 base examples and 480 model--example observations at the
bottleneck. Each metric update is divided by the absolute value of that metric
before the update, so the reported changes are relative rather than raw rank
units. Attention and MLP contributions have opposite signs in all six settings.

\begin{table}[htbp]
\centering
\small
\begin{tabular}{lrr}
\toprule
Relative update & Mean & Bootstrap 95\% CI\\
\midrule
Attention: effective rank & 0.0536 & [0.0131,0.1064]\\
MLP: effective rank & -0.0664 & [-0.1289,-0.0171]\\
Attention: cross/self & -0.0400 & [-0.0788,-0.0072]\\
MLP: cross/self & 0.0254 & [0.0066,0.0472]\\
\bottomrule
\end{tabular}
\caption{\textbf{Opposing attention and MLP contributions at the bottleneck.}
The sign of each update is replicated in all six model--dataset settings.
These measurements resolve attention and MLP contributions to the residual
representation at the sublayer level.}
\label{tab:sublayers}
\end{table}

\begin{figure}[htbp]
\centering
\includegraphics[width=\textwidth]{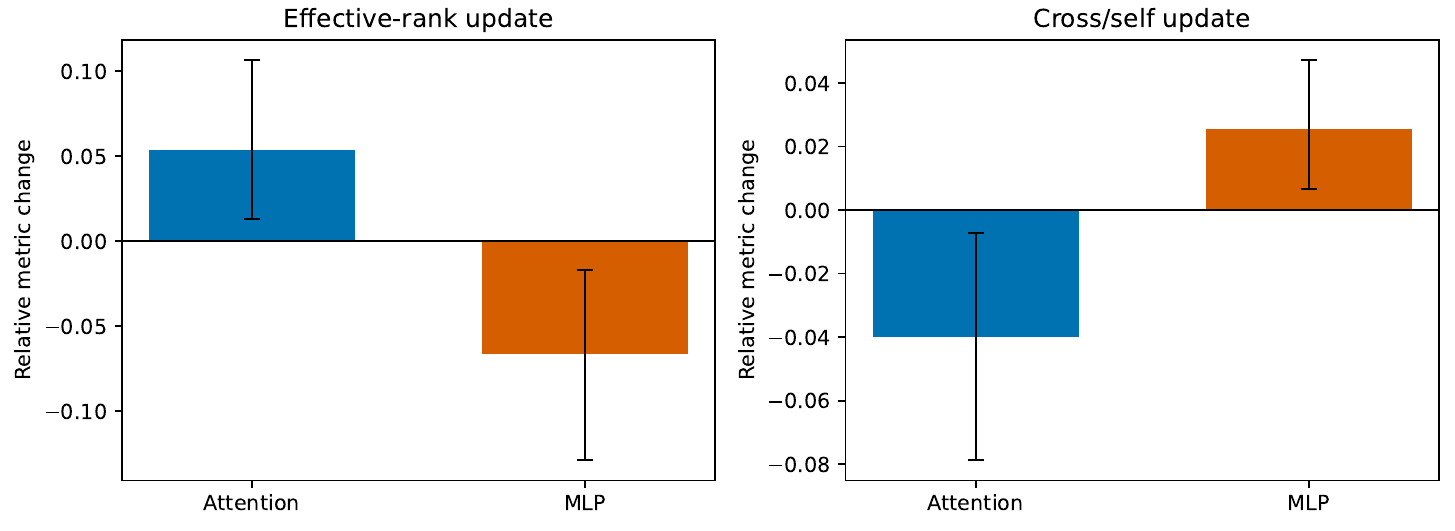}
\caption{\textbf{Scale-normalized sublayer changes.}
Rank and cross/self updates are divided by their pre-update magnitudes.
Pythia-2.8B and Qwen2.5-7B on PIQA, HellaSwag, and ARC-Challenge
show attention expansion and MLP compression, with opposing block-ratio updates.}
\label{fig:sublayers}
\end{figure}
\par\endgroup
\FloatBarrier
\Needspace{8\baselineskip}
\section{Representation Analyses and Model Diagnostics}
\label{sec:appendix_diagnostics}
The matched comparisons test spectral recovery, predictive use,
and candidate localization separately. The model-ranking analysis uses its
own feature set and validation protocol.

\par\begingroup
\Needspace{5\baselineskip}
\subsection{Existing Representation Analyses}
\label{sec:baseline_details}

This comparison tests which aspects of the covariance analysis can be recovered
from scalar spectra, normalized maps, and cross-example representation similarity.
The per-prompt study uses the canonical 14 Pythia, SmolLM, Qwen, and Mistral
models on PIQA, BoolQ, HellaSwag, ARC-Easy, ARC-Challenge, COPA, WinoGrande,
OpenBookQA, CommonsenseQA, Social IQa, SciQ, and LogiQA: up to 120 prompts per
model--dataset pair, capped at 100 for COPA, giving 19,880 observations.

\paragraph{Representations and feature budget.}
Covariance uses the centered token map defined in \Cref{sec:covariance_definition}.
Centered cosine normalizes each centered token state before forming the map.
Uncentered Gram uses $HH^\top/D$. Direct hidden-state rank summarizes the hidden
state spectrum without retaining token-pair blocks. Centered Gram equals $DC$
under the same token centering; counting it as a separate scale-invariant
baseline would duplicate covariance.

For each per-prompt representation, the four features are the minimum spectral
summary, its relative depth, its rise to the final layer, and final cross/self
where the map supports candidate blocks. For direct spectral summaries without
candidate blocks, the final top-mode fraction replaces cross/self. Features are
averaged over the same model--dataset collections. Capability regression includes
log model size, fits standardization within the training fold, and holds out one
model. Task-family classification uses standardized logistic regression and
holds out a complete model, preventing model identity from appearing in both
training and test data.

CKA~\citep{kornblith2019cka} and SVCCA~\citep{raghu2017svcca} compare layer representations with final representations across
segment-pooled examples, retaining their conventional cross-example interpretation.
Their four summaries are area under the similarity trajectory, the depth of its
largest rise, the change from initial to final similarity, and the change
from early to late similarity.
They do not produce per-prompt candidate maps.
Table~\ref{tab:baselines} reports the matched predictive comparisons.

\begin{table}[htbp]
\centering
\small
\begin{tabular}{lrrr}
\toprule
Method & Capability $R^2$ & Task macro-F1 & Candidate localization\\
\midrule
Direct hidden-state rank & \textbf{0.796} & 0.209 & 31.3\% (chance)\\
Centered cosine & 0.683 & \textbf{0.393} & 85.8\%\\
Token covariance & 0.680 & 0.245 & 83.3\%\\
Linear CKA & 0.633 & 0.146 & n/a\\
SVCCA & 0.631 & 0.195 & n/a\\
Uncentered Gram & 0.544 & 0.235 & \textbf{86.7\%}\\
\bottomrule
\end{tabular}
\caption{\textbf{Different diagnostics answer different questions.}
Capability prediction uses model size plus four features in leave-one-model-out
regression; task-family macro-F1 also holds out a complete model.
Candidate localization identifies the edited option in 960 paired prompts
from Pythia-2.8B and Qwen2.5-7B on PIQA, HellaSwag, and ARC-Challenge.
Higher values are better. Scalar rank has no candidate coordinates, so its
entry is the option-count-dependent chance comparator; CKA/SVCCA do not
produce per-prompt candidate maps.
Appendix~\ref{sec:baseline_details} specifies all features and protocols.}
\label{tab:baselines}
\label{tab:representation_baselines_summary}
\end{table}

Direct hidden-state rank recovers the covariance minimum within one sampled
depth on 96.84\% of prompts, with a mean absolute relative-depth difference
of 0.00865. The normalization-dependent differences in
interior-minimum rates are reported in Table~\ref{tab:normalization}.
These comparisons use four features and hold out a complete model.
The separate model-ranking experiment below uses a different feature set.

\begin{figure}[htbp]
\centering
\includegraphics[width=\textwidth]{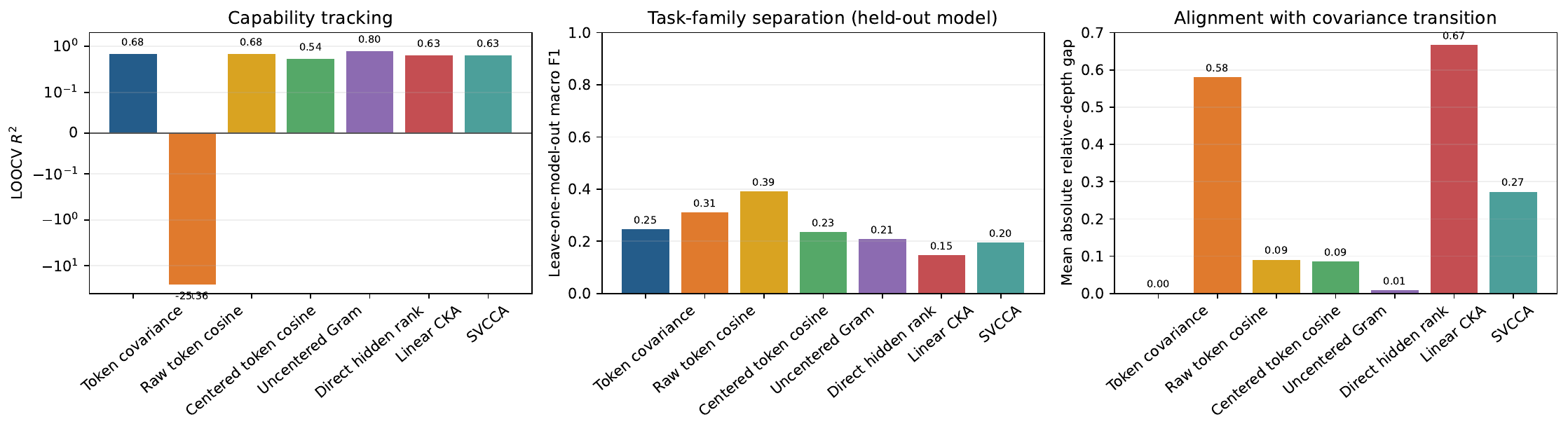}
\caption{\textbf{Existing baselines reproduce different parts of the diagnostic.}
Direct rank recovers the spectral transition and gives the best capability
prediction in the matched comparison; centered cosine gives the best task-family
transfer. Map-valued methods additionally retain candidate location. The
comparison protocols and four-feature summaries are stated in
\Cref{sec:baseline_details}.}
\label{fig:established_baselines}
\label{fig:cov_vs_simple}
\end{figure}
\par\endgroup

\FloatBarrier
\par\begingroup
\Needspace{5\baselineskip}
\subsection{Changed-Candidate Localization}
Pythia-2.8B and Qwen2.5-7B are evaluated on PIQA, HellaSwag, and
ARC-Challenge. We use 60 paired examples per valid model--dataset--edit condition:
irrelevant-distractor and correct-answer replacements for all three datasets,
plus duplicate-wrong edits for the four-choice datasets. This gives 960 paired
cases. The localization rule identifies the candidate whose map block changes
most. As shown in Table~\ref{tab:baselines} and
Figure~\ref{fig:candidate_localization}, all three token-position
maps identify the edited candidate more often than chance. Scalar rank
has no candidate coordinates, so it is assigned the option-count-dependent
chance comparator rather than a learned localization rule.

\begin{figure}[htbp]
\centering
\includegraphics[width=0.85\textwidth]{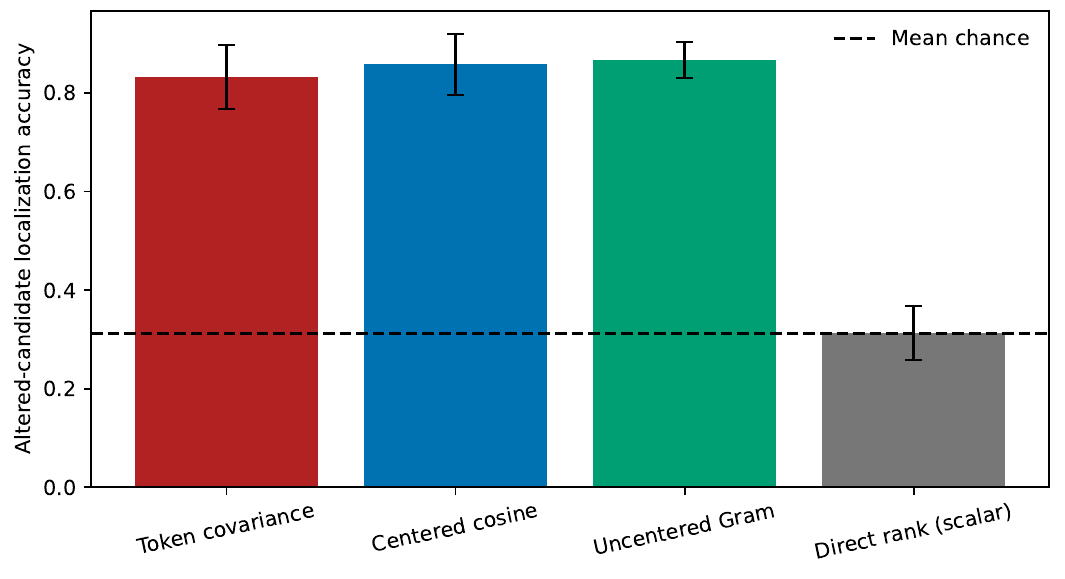}
\caption{\textbf{Token-position maps identify which candidate was changed.}
The 960 paired manipulations use Pythia-2.8B and Qwen2.5-7B on PIQA,
HellaSwag, and ARC-Challenge. The scalar-rank comparator is chance, since it
does not contain candidate coordinates. Higher localization accuracy is better.}
\label{fig:candidate_localization}
\end{figure}
\par\endgroup

\FloatBarrier
\Needspace{5\baselineskip}
\subsection{Capability and Model Ranking}

\begin{figure}[htbp]
\centering
\includegraphics[width=\textwidth]{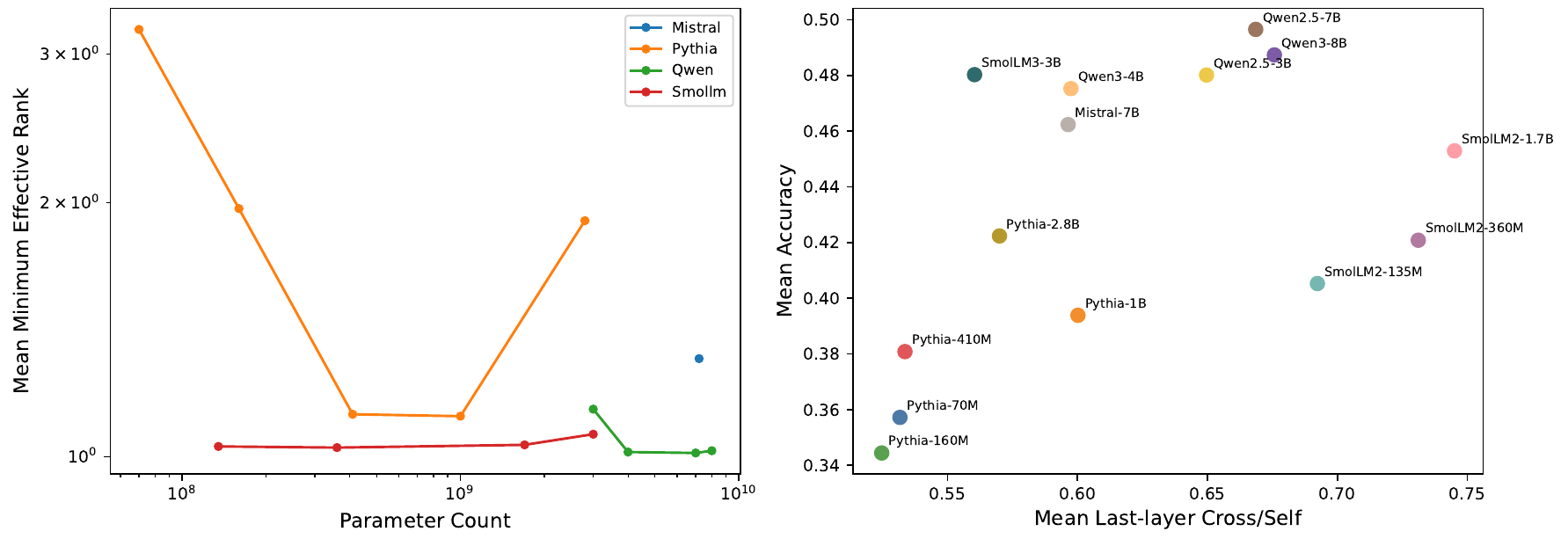}
\caption{\textbf{Supporting model-level analyses.} Left: minimum-rank scaling is heterogeneous across families. Right: late-layer option-cross interaction tracks model capability more reliably than minimum rank alone, motivating the use of late comparison summaries in the diagnostic analyses.}
\label{fig:capability}
\end{figure}

\paragraph{Capability-linked geometry and family-wise scaling.}
Figure~\ref{fig:capability} summarizes model-level diagnostic analyses. Minimum-rank
scaling is heterogeneous across families, while late-layer option-cross interaction tracks
model capability more reliably than minimum rank alone.

\begin{figure}[htbp]
\centering
\includegraphics[width=0.90\textwidth]{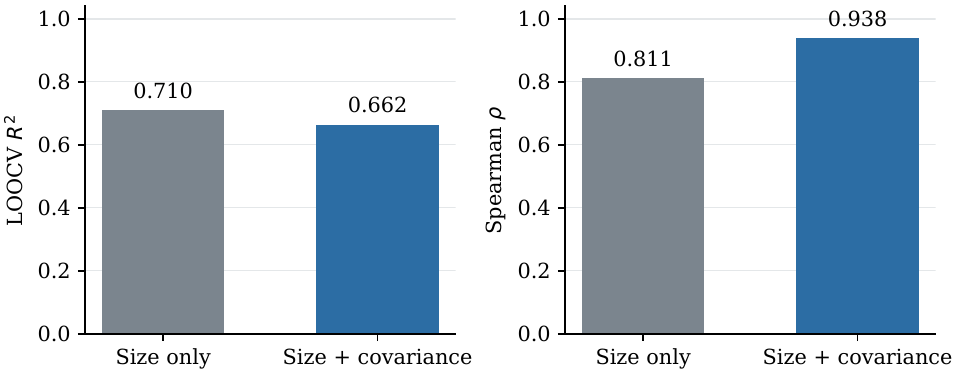}
\caption{\textbf{Model ranking with covariance summaries.}
Both predictors use the same 14-model pool and held-out-model predictions
of mean benchmark accuracy. Left: leave-one-model-out regression $R^2$;
right: Spearman correlation between those predictions and observed accuracy.
Size plus covariance improves ordering and selects the strongest model,
while size alone retains the higher regression fit.}
\label{fig:checkpoint_ranking}
\end{figure}

\paragraph{Checkpoint and model ranking diagnostic.}
Adding covariance summaries to model size improves ranking:
Spearman correlation with mean benchmark accuracy increases from 0.811
to 0.938, and the strongest model is selected. Size alone retains the
higher leave-one-model-out regression fit, as shown in
Figure~\ref{fig:checkpoint_ranking}. The ranking experiment and the matched
four-feature comparison in Table~\ref{tab:baselines} use
different feature sets and should not be compared through their regression
scores.

Figure~\ref{fig:checkpoint_ranking} shows the ranking comparison.

The selected model is Qwen2.5-7B. Selection regret, defined
as the accuracy gap between the best model and the selected model,
decreases from 0.0342 for size alone to zero with covariance.

\FloatBarrier
\Needspace{8\baselineskip}
\section{Generalization across Formats, Languages, and Tasks}
\label{sec:appendix_generalization}
We test the geometric trajectory beyond base-model English
selection prompts, distinguishing its persistence from the quality of generated answers.

\par\begingroup
\label{sec:generalization_details}

\begin{table}[htbp]
\centering
\small
\begin{tabular}{p{0.16\textwidth}p{0.64\textwidth}c}
\toprule
Evaluation & Models, tasks, and examples & Settings\\
\midrule
Official chat &
Qwen2.5-7B-Instruct and Mistral-7B-Instruct-v0.3; PIQA, BoolQ,
\mbox{HellaSwag}, ARC-Challenge; 200 per setting & 8/8\\
Multilingual &
Qwen2.5-7B base and Instruct; aligned XCOPA in five languages;
Estonian, Indonesian, Swahili, Tamil, Chinese; 100 shared items & 10/10\\
Professional\newline domains &
Both instruction-tuned models; MMLU professional law, clinical knowledge,
abstract algebra; 100 per subject & 6/6\\
Generation &
Pythia-2.8B, Qwen2.5-7B, Mistral-7B-v0.3;
3,200 responses across 20 model--task settings and seven prefixes & 140/140\\
\bottomrule
\end{tabular}
\caption{\textbf{Compression and re-expansion extend across templates, languages,
and domains.} The final column counts model--task or model--language settings
with an interior minimum and positive late re-expansion; generation counts
model--task--prefix settings. The models, tasks, and prompt formats are
described below. Adjusted pooled
multilingual re-expansion is 28.60, with bootstrap 95\% CI [28.11,29.10].}
\label{tab:generalization}
\label{tab:generalization_summary}
\end{table}

\Needspace{5\baselineskip}
\subsection{Instruction-Tuned Models}
Qwen2.5-7B-Instruct and Mistral-7B-Instruct-v0.3 use the official
chat template provided by each tokenizer on PIQA, BoolQ, HellaSwag, and ARC-Challenge, with 200 examples
per model--dataset setting. The same examples are evaluated in the plain
unified template. All eight chat settings retain an interior bottleneck and
positive re-expansion; bottleneck depth is unchanged on 1,600 paired examples.

\Needspace{5\baselineskip}
\subsection{Aligned Multilingual Evaluation}
The multilingual study uses Qwen2.5-7B and Qwen2.5-7B-Instruct on the same
100 XCOPA~\citep{ponti2020xcopa} item IDs in Estonian, Indonesian, Swahili, Tamil, and Chinese.
All ten model--language settings retain the trajectory at relative depth 0.1111.
Adjustment includes token count, character count, tokens per character, and
option-length imbalance, with bootstrap resampling over aligned item IDs.
Adjusted pooled re-expansion remains positive. Raw magnitudes differ across
languages, most visibly for Tamil; bottleneck location remains stable while
re-expansion magnitude varies with the tokenization regime.

\Needspace{5\baselineskip}
\subsection{Professional Domains}
Both instruction-tuned models use 100 examples each from MMLU~\citep{hendrycks2021mmlu} professional law,
clinical knowledge, and abstract algebra. All six model--subject settings
retain an interior minimum and positive re-expansion.

\Needspace{5\baselineskip}
\subsection{Open-Ended Generation and Replay}
The online study uses base checkpoints
\nolinkurl{EleutherAI/pythia-2.8b-deduped},
\nolinkurl{Qwen/Qwen2.5-7B}, and \nolinkurl{mistralai/Mistral-7B-v0.3}.
Each generates 200 responses per option-removed PIQA, BoolQ, HellaSwag, and
ARC-Challenge task (2,400 responses). Base Qwen and Mistral additionally
generate 100 responses on GSM8K~\citep{cobbe2021verifiers},
TriviaQA~\citep{joshi2017triviaqa}, MBPP~\citep{austin2021program},
and QASPER~\citep{dasigi2021qasper} (800 responses).

We preserve the question/context and remove the option list. For each realized
response, we replay question-only, 25\%, 50\%, 75\%, pre-answer,
first-answer-token, and complete prefixes through every transformer layer.
The map is segmented by question/context, continuation, and answer tokens.
Input and replay budgets are fixed at 1,024 and 1,536 tokens respectively;
QASPER context is deterministically truncated within the input budget.
Generation is deterministic, with task-specific new-token caps of 96 for
option-removed tasks, 128 for GSM8K, 64 for TriviaQA, 192 for MBPP, and 128
for QASPER. Geometry is analyzed without filtering on correct answers.

The answer onset follows the last ``Final answer:''
marker when it is present; otherwise it is set deterministically at 80\%
of the continuation. QASPER uses truncated context within the 1,024-token input limit.

For each prefix, we measure the increase from minimum to final-layer
effective rank. The 140/140 count refers to 20 model--task settings
times seven prefixes, each retaining interior compression and re-expansion.
The separate 19/20 count compares completion with question-only re-expansion.
These are different endpoints and do not imply monotonic improvement for every
response.
A model--task--prefix setting meets our criterion when the lower
95\% confidence bounds for both the interior-minimum rate and the
positive-re-expansion rate exceed 0.5. This criterion concerns the
responses within each setting, not 140 independent model
evaluations.

\begin{figure}[htbp]
\centering
\includegraphics[width=\textwidth]{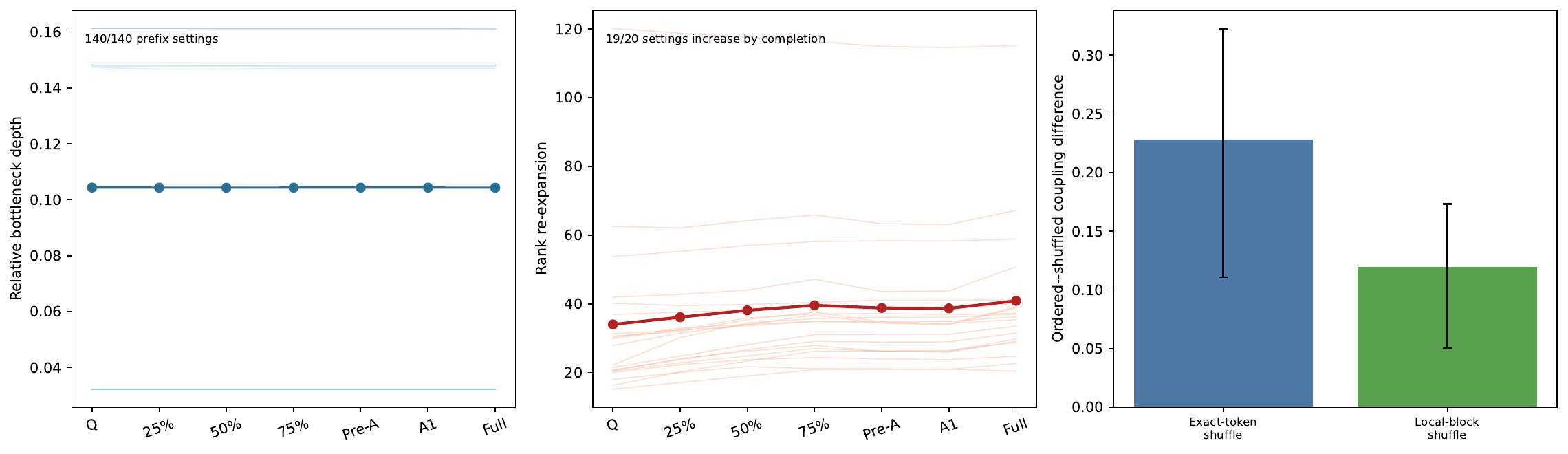}
\caption{\textbf{The trajectory persists as generation proceeds.}
The 3,200-response study covers three base models on four option-removed tasks
and Qwen/Mistral on GSM8K, TriviaQA, MBPP, and QASPER.
Left: bottleneck depth at seven prefixes. Center: the increase from minimum
to final effective rank for each prefix. Right: continuation controls. All 140 model--task--prefix
settings retain compression and re-expansion; completion exceeds question-only
re-expansion in 19 of 20 model--task settings.
Intervals are hierarchical bootstrap 95\% confidence intervals.
Appendix~\ref{sec:generalization_details} defines the distinct
rank and block-coupling endpoints.}
\label{fig:online}
\end{figure}

\Needspace{5\baselineskip}
\subsection{Continuation-Order Controls}
Exact-token shuffling reorders continuation tokens before the answer while
retaining their multiset and length. Local-block shuffling reorders contiguous
four-token blocks, preserving order inside each block. Answer tokens remain
fixed. For these controls we measure the bottleneck-to-final rise in
continuation--answer absolute covariance, normalized by the mean absolute
covariance of the full map. Ordered continuations have a larger normalized
coupling rise than either shuffle, as shown in Figure~\ref{fig:online}.
This endpoint measures block coupling, not effective-rank units. It tests
sensitivity to continuation order beyond sequence growth, not answer correctness.

\begin{table}[htbp]
\centering
\small
\begin{tabular}{lrr}
\toprule
Continuation control & Ordered--shuffled difference & Bootstrap 95\% CI\\
\midrule
Exact-token shuffle & 0.228 & [0.111,0.322]\\
Local four-token-block shuffle & 0.120 & [0.051,0.173]\\
\bottomrule
\end{tabular}
\caption{\textbf{Continuation order contributes beyond length and token content.}
The endpoint is the bottleneck-to-final rise in continuation--answer absolute
covariance divided by the mean absolute covariance of the map. Answer tokens are
fixed in both controls. The 3,200 responses cover 20 model--task settings;
intervals use hierarchical bootstrap resampling, not effective-rank units.}
\label{tab:continuation_order_effects}
\end{table}

Professional-domain and generation quality varies across tasks and models.
The extension concerns the geometric trajectory and ordered continuation
structure, with answer quality evaluated separately.
For generation without listed candidates, we measure
compression, re-expansion, and sensitivity to continuation order. These
endpoints extend the geometric profile observed in structured selection;
identifying answer comparison during generation requires an additional
candidate-level behavioral test.
\par\endgroup
\FloatBarrier
\Needspace{8\baselineskip}
\section{Practical and Behavioral Diagnostics}
\label{sec:appendix_side_analyses}
These analyses test whether geometry supports selective
computation or predicts correctness, rather than assuming that diagnostic
structure is sufficient for reliable control.

\Needspace{5\baselineskip}
\subsection{Geometry-Augmented Selective Reranking}
We test whether geometry helps decide when to replace direct multiple-choice
answering with pairwise reranking. Geometry-Augmented Selective Reranking
(GASR) is compared with a matched confidence-only controller and simple
threshold policies in Table~\ref{tab:gasr_summary}. The pooled mean accuracy
is slightly higher for GASR, but fixed-budget bootstrap intervals include
zero and the advantage does not hold in a majority of model--dataset
settings. The result therefore does not establish reliable online control.

\begin{table}[htbp]
\centering
\small
\begin{tabular}{lcc}
\toprule
\textbf{Policy} & \textbf{Pooled mean budgeted accuracy} & \textbf{Role} \\
\midrule
GASR & 0.4731 & Geometry-augmented two-stage controller \\
TwoStageConf & 0.4690 & Matched confidence-only controller \\
MarginThreshold & 0.4430 & Simple threshold baseline \\
EntropyThreshold & 0.4435 & Simple threshold baseline \\
AlwaysDirect & 0.4325 & No reranking baseline \\
\bottomrule
\end{tabular}
\caption{\textbf{Selective-reranking results.} GASR improves pooled mean
budgeted accuracy over the matched confidence-only controller and simple threshold
baselines, but the fixed-budget bootstrap gaps remain statistically inconclusive.}
\label{tab:gasr_summary}
\end{table}

\Needspace{5\baselineskip}
\subsection{Adaptive Routing}
Geometry-aware routing is a plausible downstream application, but in the current
experiments it does not outperform simple margin-based baselines at matched budgets. The
selective-reranking benchmark likewise gives a small pooled
improvement, but fixed-budget gains are inconsistent across model--dataset settings.

\begin{figure}[htbp]
\centering
\includegraphics[width=\textwidth]{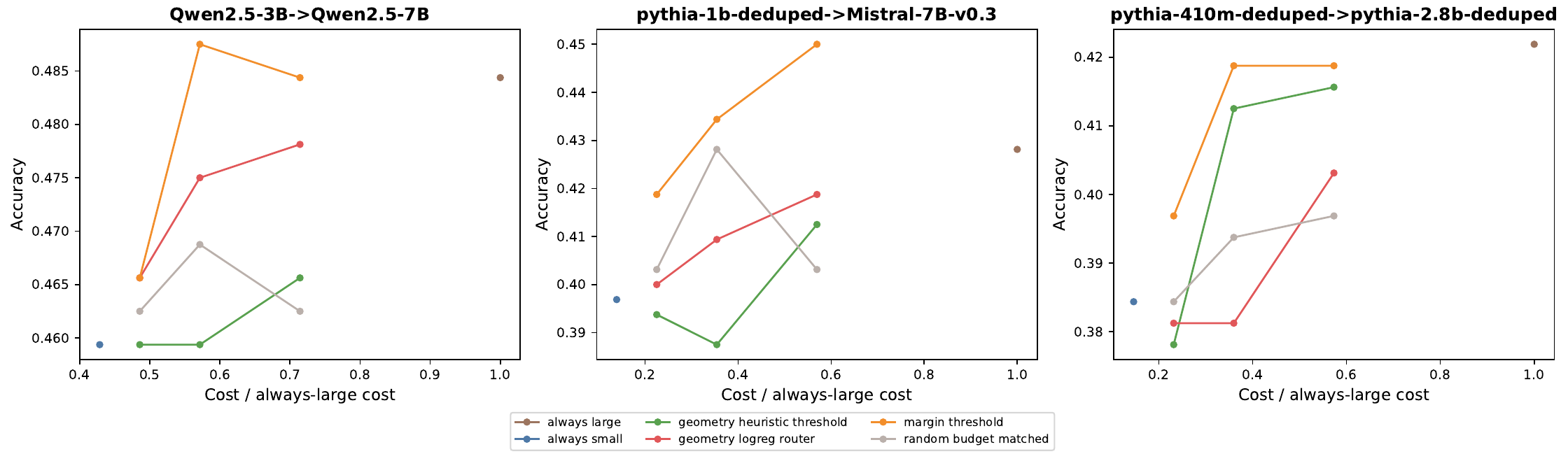}
\caption{\textbf{Adaptive routing at matched budgets.} Geometry-aware routing
does not consistently improve on margin-based thresholding. The curves
compare the accuracy--cost tradeoff rather than establish a deployment advantage.}
\label{fig:routing}
\end{figure}

Figure~\ref{fig:routing} shows the routing Pareto curves.

\Needspace{5\baselineskip}
\subsection{Correctness Prediction}
\begin{figure}[htbp]
\centering
\includegraphics[width=\textwidth]{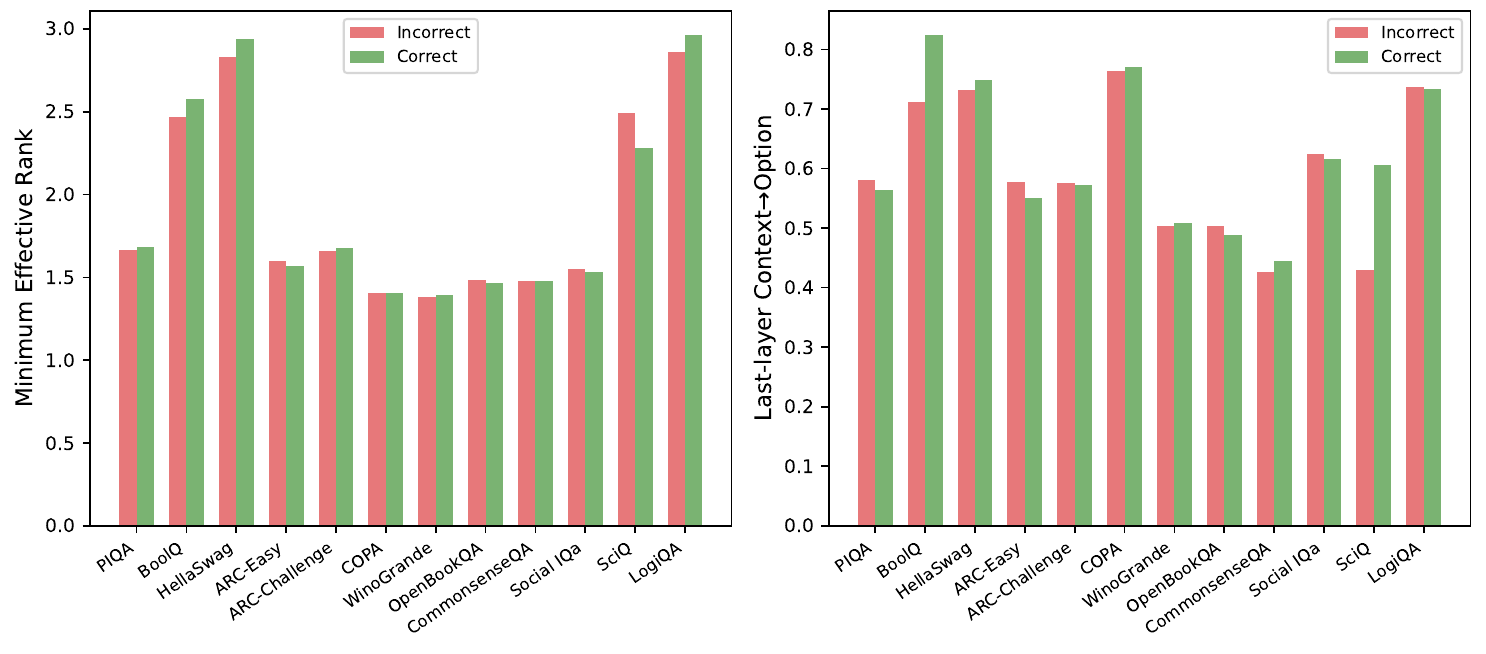}
\caption{\textbf{Correctness-conditioned geometry.} For Pythia-2.8B, geometry contains some per-example success signal, but the association varies across settings.}
\label{fig:correct}
\end{figure}

We test whether the geometric summaries predict correctness for individual
examples. As shown in Figure~\ref{fig:correct}, logistic regression separates
correct and incorrect responses in some settings, but the association does
not repeat consistently across the representative larger models.

\end{document}